\documentclass{article}

\usepackage[main, final]{neurips_2026}

\usepackage[utf8]{inputenc} 
\usepackage[T1]{fontenc}    
\usepackage{url}            
\usepackage{booktabs}       
\usepackage{amsfonts}       
\usepackage{nicefrac}       
\usepackage{microtype}      
\usepackage{xcolor}         
\usepackage{url}
\usepackage[colorlinks=true,
            linkcolor=purple,
            citecolor=blue,
            urlcolor=purple]{hyperref}
\usepackage{microtype}
\usepackage{graphicx}
\usepackage{subcaption}
\usepackage{booktabs,multicol,multirow} 
    \usepackage{longtable}
\usepackage{amsmath}
\usepackage{amssymb}
\usepackage{algorithm, algorithmic}
\usepackage{mathtools}
\usepackage{amsthm}
\usepackage{tikz}
\usepackage{wrapfig}
\usepackage{xcolor,lipsum,enumitem}
\usepackage[capitalize,noabbrev]{cleveref}

\theoremstyle{plain}

\theoremstyle{definition}

\theoremstyle{remark}

\newcommand{\std}[1]{\scriptsize{$\pm$#1}}

\title{Graph Label Alignment:\\ A Diagnostic Atlas for Graph Classification}
\author{%
  Neelam Akula\\ 
  Department of Mathematical Sciences\\
  University of Texas at Dallas\\
  Richardson, TX 75080\\
  \texttt{neelam.akula@utdallas.edu}
  \And
  Varun Shiralkar\\
  Department of Computer Science\\
  University of Texas at Dallas\\
  Richardson, TX 75080\\
  \texttt{varun.shiralkar@utdallas.edu}
  \And
  Murat Kantarcioglu\\
  Department of Computer Science\\
  Virginia Tech\\
  Blacksburg, VA 24061\\
  \texttt{muratk@vt.edu}
  \And
  Baris Coskunuzer\\
  Department of Mathematical Sciences\\
  University of Texas at Dallas\\
  Richardson, TX 75080\\
  \texttt{coskunuz@utdallas.edu}
}

\begin{document}

\maketitle

\begin{abstract}
Graph classification benchmarks vary widely in difficulty: some are solved by simple pooled baselines, while others remain challenging for GNNs and graph transformers. We take a data-centric view and ask whether benchmark success is driven by genuine structure--feature reasoning or by label signal already accessible from restricted marginal views. We introduce \emph{graph label alignment}, a lightweight diagnostic that measures label accessibility through two supervised probes computed without training a graph encoder: \emph{structural label alignment} (SLA), based on permutation-invariant topological and spectral summaries, and \emph{feature label alignment} (FLA), based on pooled node features with edges removed. Across real graph classification benchmarks and multiple pooled backbones, the resulting SLA--FLA plane strongly organizes test accuracy, with low-alignment datasets forming a consistent regime where standard pooled pipelines struggle. To probe what lies beyond marginal accessibility, we introduce controlled interaction-dominated stress tests, including heterophilic ego-graphs and synthetic \emph{Twins} tasks whose structural and feature marginals are matched across classes but whose role--feature couplings differ. These results show that high alignment explains much of easy benchmark performance, while low alignment reveals regimes where success requires preserving structure--feature correspondence beyond standard pooled representations.
\end{abstract}

\section{Introduction}
Strong performance on a graph classification benchmark can mean two very different things: a model may exploit relational structure, or it may exploit label signal already accessible from coarse marginal views of the data. This ambiguity is increasingly important because graph classification benchmarks vary widely in difficulty: many are handled well by simple pooled baselines, while others challenge expressive GNNs and graph transformers~\citep{errica2020fair,dwivedi2023benchmarking}. This motivates a data-centric question: \emph{Where does the label signal live in a graph classification dataset, and which parts of it are accessible to standard pooled graph models?}

We address this question through a supervised restricted-view diagnostic for graph classification. The goal is to separate marginal label accessibility from interaction-dependent signal: high marginal accessibility can explain why many benchmarks are easy for standard pooled pipelines, while low marginal accessibility highlights regimes where success requires preserving structure--feature correspondence beyond coarse pooled summaries. If labels are predictable from structure alone or from the multiset of node features alone, then strong accuracy may largely reflect this accessibility rather than richer relational reasoning~\citep{li2024rethinking,coupette2025rings}. Conversely, when both restricted views are weak, remaining signal must come from higher-order structure--feature interactions, information not captured by the reference probes, or noise. We instantiate this low-marginal regime through controlled Twins tasks, where structural and feature shortcuts are eliminated by design.

\paragraph{Graph label alignment.} Given a dataset $\mathcal{D}=\{(G_i,X_i,y_i)\}_{i=1}^N$, we introduce two probe-based scores that quantify marginal label accessibility without training a graph encoder. \textit{Structural label alignment (SLA)} measures label predictability from a fixed, permutation-invariant structural descriptor of $G_i$ alone, based on topological and spectral summaries. \textit{Feature label alignment (FLA)} measures label predictability when edges are discarded and only the multiset of node features in $X_i$ is retained. Together, $(\widehat{\mathrm{SLA}},\widehat{\mathrm{FLA}})$ place datasets on an interpretable \emph{graph label alignment plane}, enabling benchmark comparison and helping identify when performance is likely driven by marginal accessibility versus structure--feature interaction.

\paragraph{A plane that organizes pooled-model performance.} Across real benchmarks and pooled graph classifiers, including GCN, GraphSAGE, GAT, and GPS, datasets with higher SLA or FLA tend to yield higher accuracy, while low-alignment datasets cluster in regimes where pooled backbones generally perform worse. Architecture choice shifts absolute performance, but does not reliably remove this difficulty. We quantify the trend using Pearson correlations between alignment scores and test accuracy, and use $\mathrm{AlignAvg}=(\mathrm{SLA}+\mathrm{FLA})/2$ as a compact dataset ordering.

\paragraph{Controlled stress tests expose what the alignment plane misses.} Real datasets confound domain, size, feature type, and class structure, so we complement them with controlled settings. A synthetic sweep independently dials structural and feature separation, validating that accuracy contours follow the alignment geometry. Heterophilic ego-graph classification provides realistic low-marginal tasks where pooled message-passing baselines achieve low accuracy, typically $\sim$25--45\%, and performance is sensitive to hop radius. Most sharply, in balanced binary and three-class \emph{Twins} tasks, classes have identical structural and feature marginals but distinct role--feature couplings; pooled GNNs and GPS remain near random guessing. These results suggest a correspondence bottleneck in standard pooled graph-classification pipelines: when label signal resides in global role--feature coupling, simply changing the encoder may not preserve the information needed for discrimination.

\paragraph{Contributions.}
\begin{itemize}[leftmargin=*]
	\item We propose \textbf{graph label alignment}, a label-aware diagnostic for pooled graph classification, and define probe-based \textbf{SLA} and \textbf{FLA} scores for structure-only and feature-only restricted views.
	\item We show that pooled GNN and graph transformer performance on real benchmarks is strongly organized by the \textbf{SLA--FLA plane}, supporting $\mathrm{AlignAvg}$ as a compact dataset ordering.
	\item We provide \textbf{backbone-specific alignment predictors} and a practical \textbf{decision map} (Appendix~\ref{app:gla-recipe}) that translate SLA/FLA regimes into backbone selection guidance.
	\item We introduce controlled \textbf{synthetic alignment sweeps} and \textbf{interaction-dominated stress tests}, including heterophilic ego-graphs and \emph{Twins}, showing that low-marginal regimes can expose a failure mode of standard pooled pipelines under role--feature coupling.
\end{itemize}

\section{Background}\label{sec:background}
\subsection{Related Work}
\paragraph{Graph classification models and benchmarks.} Graph classification is a core task in graph machine learning with applications in chemistry, biology, and social systems. Most methods learn node representations and aggregate them into graph-level embeddings, typically using message-passing GNNs with global pooling, or attention-based graph transformers with structural encodings. Evaluation is commonly performed on benchmark suites such as TU and OGB graph-level datasets \citep{morris2020tudataset,hu2020ogb}. Prior work has emphasized that graph classification results can be highly sensitive to experimental protocols and that carefully tuned simple baselines can be competitive on several benchmarks \citep{errica2020fair}.

\paragraph{Data-centric evaluation of graph classification datasets.} A growing line of work asks whether popular benchmarks effectively test graph reasoning or whether they can be solved by simpler cues. \cite{li2024rethinking} revisit graph classification benchmarks and propose effectiveness metrics based on performance gaps between graph-aware methods and simpler structure-only or attribute-only baselines, together with correlation-controllable synthetic datasets. \cite{coupette2025rings} propose RINGS, a mode-perturbation framework that evaluates datasets through model sensitivity to structure and feature perturbations, using measures such as performance separability and mode complementarity. These works assess benchmark quality via performance gaps and perturbation sensitivity.

In contrast, our goal is not to curate benchmarks, but to provide a simple \emph{label-centric} view of where predictive signal is accessible under restricted information views. We introduce two probe-based scores: \emph{structural label alignment} (SLA), measuring label predictability from permutation-invariant structural summaries alone, and \emph{feature label alignment} (FLA), measuring label predictability from node-feature pooling summaries when edges are discarded. Together, they define an interpretable alignment plane that separates structure-accessible, feature-accessible, and low-marginal-accessibility regimes without training full graph encoders.

\paragraph{Alignment between structure, features, and labels.} Several works study alignment between graph structure and node features, typically for node-level learning, and use spectral criteria to guide rewiring or denoising. For example, \cite{linkerhagner2025joint} propose jointly rewiring the graph and denoising features by aligning leading spectral spaces of graph and feature matrices. Our setting differs in that we study \emph{graph-level} classification and define alignment directly in terms of \emph{label predictability} under restricted views, rather than structure-feature agreement itself. We use the term \emph{alignment} strictly in this predictive sense.

We further use synthetic and ego-graph datasets to study regimes where labels are accessible from structure, accessible from features, or suppressed under both marginal views. The low-marginal regime is especially useful diagnostically: it can indicate noise, probe underfitting, or genuinely interaction-dependent signal. Our controlled Twins construction isolates the last case by matching structural and feature marginals across classes while varying only role-feature coupling. This lets us compare datasets and architectures on a common plane while making explicit what the alignment scores do and do not establish.

\subsection{Structural \& Feature Probes}
We quantify label accessibility under two restricted views: structure-only and feature-only. For each view, we construct lightweight, permutation-invariant summary representations that serve as standardized probes across datasets. These probes are intentionally simple and are used to define controlled diagnostics for SLA and FLA (Section~\ref{sec:method}); they are not designed to maximize predictive performance. Implementation details are provided in Appendix~\ref{app:summary_vectors}.

\paragraph{Structural summaries.} We represent each graph using a combination of topological, spectral, and statistical descriptors. For topology, we compute Betti-profile summaries derived from persistent homology over a filtration (e.g., node degree) \citep{dey2022computational,coskunuzer2024topological} (Appendix~\ref{sec:PH}). For spectral structure, we include a small set of extremal eigenvalues of the normalized Laplacian and adjacency matrices. Together, these features provide compact, permutation-invariant summaries of graph structure (see Appendix~\ref{app:summary_vectors}).

\paragraph{Feature-only summaries.} For the feature-only view, we discard edges and treat node features as an unordered multiset. Each graph is represented by a permutation-invariant pooled descriptor; in our reference setting, we use the feature centroid. Alternative summaries are evaluated in Appendix~\ref{app:summary_vectors}.

\section{Graph--Label Alignment Framework}\label{sec:method}
\subsection{Problem Setup}\label{subsec:setup}
We consider a graph classification dataset $\mathcal{D}=\{(G_i,X_i,y_i)\}_{i=1}^N$, where $G_i=(V_i,E_i)$ is a graph, $X_i\in\mathbb{R}^{|V_i|\times d}$ are node features, and $y_i\in\{1,\dots,C\}$ is a graph label. We view each sample through two restricted information views: \textbf{structure-only} (discard $X_i$) and \textbf{feature-only} (discard $E_i$), and ask how much label signal is accessible from each view via standardized, lightweight probe predictors. Throughout, we use \emph{alignment} strictly to mean \emph{label predictability under a restricted view}, rather than agreement between structure and features (see \Cref{sec:background}).

\subsection{Restricted-View Summary Vectors}\label{subsec:summary_vectors}
To quantify label signal under each view, we map each graph to a fixed-dimensional summary vector. These summaries are deliberately simple and permutation-invariant: they are not intended to maximize accuracy, but to provide controlled, comparable diagnostics across datasets.

\paragraph{Structural view.} Node features are discarded and each graph $G$ is represented by a \emph{Structural Summary Vector} $\phi_{\mathrm{struct}}(G)\in\mathbb{R}^{48}$ that concatenates three families of permutation-invariant descriptors: a degree-filtration Betti profile (PH, 20 dims), a compact spectral signature from the normalized Laplacian and adjacency matrix (8 dims), and global graph statistics including size, density, degree quantiles, and clustering moments (20 dims). The vector uses no attribute information; zero-padding is applied when fewer than four nonzero eigenvalues are available. Full construction details are in Appendix~\ref{app:summary_vectors}; component-level ablations supporting their complementarity are in Appendix~\ref{app:struct-ablation}.

\paragraph{Feature-only view.} We discard edges and treat each graph as an unordered multiset of node features, and  represent each graph by a \emph{Feature Summary Vector} $\phi_{\mathrm{feat}}(G)\in\mathbb{R}^{d}$ given by the feature centroid, $\phi_{\mathrm{feat}}(F) = \nicefrac{1}{|V|}\sum_{v\in V}X_v$. Alternative feature summaries considered are reported in App.~\ref{app:summary_vectors}.

\subsection{Alignment Score Computation}\label{subsec:alignment_scores}
Given summary vectors $\phi(G_i)\in\{\phi_{\mathrm{struct}}(G_i),\phi_{\mathrm{feat}}(G_i)\}$, we estimate label accessibility using a fixed probe protocol. We train two lightweight classifiers---logistic regression and an RBF-kernel SVM---using seeded, stratified cross-validation, and define each alignment score as the average of their cross-validated accuracies.

When $\phi=\phi_{\mathrm{struct}}$, this defines the \emph{Structural Label Alignment} score $\widehat{\mathrm{SLA}}$, measuring how predictive graph structure alone is of the label. When $\phi=\phi_{\mathrm{feat}}$, it defines the \emph{Feature Label Alignment} score $\widehat{\mathrm{FLA}}$, measuring how informative the marginal node-feature distribution is when adjacency is ignored.

Both scores are operational restricted-view diagnostics, not causally isolated measures of structural or feature contributions. If node features encode structural information, as atom types do in molecular graphs where valence partly determines local topology, FLA absorbs that signal by design. Thus FLA asks ``how much label signal is accessible from features alone?'', not ``how much signal is causally feature-driven?'' Similarly, low SLA means structure is not predictive \emph{under the chosen structural summary}, not that no structural signal exists.

Probe details are in Appendix~\ref{app:align-score}. To assess whether the reference protocol underestimates accessible signal on a given dataset, Appendix~\ref{app:probe-sufficiency} introduces  a \emph{probe sufficiency self-check} that compares reference scores against an alternative five-classifier suite and flags probe-sensitive datasets.

\subsection{Using Alignment in Analysis}\label{subsec:perf_predict}
We use $(\widehat{\mathrm{SLA}},\widehat{\mathrm{FLA}})$ as a two-dimensional summary of marginal label accessibility for pooled graph classification pipelines. In \Cref{sec:results}, we visualize datasets in the SLA--FLA plane and analyze pooled-model behavior via regime-based comparisons. We also use $\mathrm{AlignAvg}=(\mathrm{SLA}+\mathrm{FLA})/2$ (when both are defined) as a one-number ordering of alignment for correlation analysis and plotting.

Two properties of this plane are worth noting upfront. First, SLA and FLA are structurally independent axes: the synthetic sweep in \Cref{sec:synthetic} confirms they decouple cleanly under controlled generation, and several real datasets (e.g., \textsc{MolHIV}, \textsc{Letter-High}, \textsc{Coil-Del}) show meaningful SLA--FLA gaps. The positive correlation observed across the broader benchmark suite is an empirical property of existing collections, not a built-in property of the framework: in many current benchmarks, datasets with high marginal accessibility under one view also tend to exhibit high marginal accessibility under the other. Second, low alignment on both axes does not automatically imply interaction-driven signal: it means labels are not accessible from either marginal view, which can arise from genuine noise, from signal in higher-order structure--feature interactions, or from probe underfitting on atypical datasets (see Appendix~\ref{app:probe-sufficiency}). The distinction matters for interpretation and is addressed directly in the controlled experiments of \Cref{sec:experiments}.

\section{Synthetic Alignment Testbeds}\label{sec:synthetic}
We use controlled synthetic graph families for two purposes. First, an \emph{alignment plane sweep} spans a range of SLA and FLA values and tests whether pooled-model performance follows the $(\widehat{\mathrm{SLA}},\widehat{\mathrm{FLA}})$ geometry under controlled variation. Second, the \emph{Twins} construction removes marginal shortcuts and isolates interaction-only signal: prediction is possible only if a model captures how node features are arranged relative to graph structure.

\subsection{Alignment Plane Sweep}\label{subsec:synth_sweep}
\paragraph{Goal.} The sweep instantiates three regimes: structure-only (high SLA, low FLA), feature-only (high FLA, low SLA), and jointly aligned (high SLA, high FLA). By varying structural and feature signal independently, we obtain datasets whose location in the alignment plane can be controlled while other factors are fixed.

\paragraph{Base graph generator.} For each graph with label $y\in\{1,\dots,K\}$, with $K=3$, structure is generated from a label-controlled mixture. Let $s\in[0,1]$ be the \emph{structural separation} parameter. With probability $s$, the graph is sampled from a class-specific generator $\mathcal{G}_y$; with probability $1-s$, it is sampled from a shared label-independent generator $\mathcal{G}_0$.

The class-specific generators are drawn from canonical random graph families such as Erd\H{o}s--R\'enyi, Barab\'asi--Albert, Watts--Strogatz, and two-block stochastic block models, with class-dependent parameters chosen to induce structural differences. The shared generator $\mathcal{G}_0$ is an Erd\H{o}s--R\'enyi model with fixed edge probability. We add label-agnostic edge rewiring noise by rewiring a fixed fraction $p_{\mathrm{rewire}}$ of edges uniformly at random, and sample graph sizes from a fixed distribution so expected node counts are matched across classes. Thus $s\approx 0$ produces structure-agnostic datasets, while $s\approx 1$ produces structure-informative datasets. Full parameter values are in Appendix~\ref{app:synthetic}.

\paragraph{Feature generator.} Node features are generated independently of graph size. For each node $v$, we sample $x_v\sim\mathcal{N}(0,\sigma^2 I)$ with fixed $\sigma^2$. Let $f\in[0,1]$ be the \emph{feature separation} parameter. With probability $f$, all nodes in a graph of label $y$ receive a class-dependent mean shift $\mu_y$; otherwise features remain label-independent noise. The class means are sampled once, normalized, and scaled by a fixed magnitude. Thus increasing $f$ makes pooled feature summaries more label-predictive, while $f\approx 0$ matches marginal feature distributions across classes.

\paragraph{Regime instantiations.} We sweep $(s,f)$ over a fixed grid to cover the SLA--FLA plane. This yields three families: \emph{Structure-only} (vary $s$ with $f=0$), \emph{Feature-only} (vary $f$ with $s=0$), and \emph{Jointly aligned} (set $s=f$). Grid size and graphs per cell are given in Appendix~\ref{app:synthetic}.

\subsection{Twins Construction}\label{subsec:twins}
\paragraph{Goal.} Twins eliminate structure-only and feature-only shortcuts. By construction, all classes share the same graph distribution and the same multiset of node feature types, so SLA and FLA are near chance. Label signal comes only from how feature types are coupled to structural roles.

\paragraph{Per-graph labeled variants.} For each base graph $G=(V,E)$ sampled from a fixed-parameter SBM~\citep{holland1983stochastic}, we generate multiple labeled variants that share the same edge set $E$ and the same balanced multiset of feature types. Each variant contains exactly $n/3$ nodes of each prototype type, so the marginal feature distribution is matched across labels. Variants differ only in the assignment of feature prototypes to nodes, forcing successful models to detect role--feature coupling rather than structure-only or feature-only signal.

\paragraph{Role scores and quantile bins.} To induce structured feature placement, we compute a scalar role score $s(v)$ for each node and partition nodes into three equal-sized bins by tertiles. We consider four scoring functions, detailed in Appendix~\ref{app:synthetic}: \textbf{Linear}, \textbf{Radial}, \textbf{PageRank}, and \textbf{Eigen}. Feature prototypes $(\mu_1,\mu_2,\mu_3)$ are then assigned to the bins in class-specific orders with additive Gaussian noise, while preserving identical prototype counts across classes.

\paragraph{Classes.} Label $0$ corresponds to a random assignment obtained by uniformly permuting prototype labels across nodes while preserving the $n/3$--$n/3$--$n/3$ counts. Labels $1,2,\dots$ correspond to structured assignments that align prototype types with role-score bins. We consider binary Twins (random vs.\ one structured assignment) and three-class Twins (random vs.\ two structured assignments). Since both structure and feature marginals are matched, any gain over chance must come from preserving global role--feature correspondence.

\section{Experiments} \label{sec:experiments}
\paragraph{Overview and evaluation protocol.} Our goal is to characterize how \emph{pooled} graph classifiers behave as a function of graph--label alignment, and to separate two questions: (i) whether marginal label accessibility (SLA/FLA) organizes performance on \emph{standard} graph benchmarks, and (ii) what fails when labels depend primarily on \emph{structure--feature interaction} that permutation-invariant readouts may wash out. We compute SLA/FLA \emph{once per dataset} using the restricted-view probes from \Cref{sec:method}; these scores are model-agnostic and require no GNN training. To answer (i), we use \textbf{general benchmarks} spanning diverse domains and naturally covering a wide range of SLA/FLA: \textbf{(1a)} real graph classification datasets, and \textbf{(1b)} a synthetic alignment-plane sweep, where we can dial structural and feature signal continuously while holding other factors fixed. To answer (ii), we use \textbf{interaction-driven stress tests} designed to suppress marginal shortcuts and stress coupling: \textbf{(2a)} heterophilic node-to-graph ego networks (Appendix~\ref{app:ego_details}), which provide realistic interaction-dominated instances, and \textbf{(2b)} synthetic \emph{Twins}, which match structure and feature marginals across classes by construction and isolate role--feature coupling as the only discriminative signal.

For real benchmarks and ego-graphs, we follow the standard dataset splits and evaluation protocols; for synthetic datasets, we use balanced classes with a 60/20/20 train/validation/test split (\Cref{app:experiments}). Our code is available at the link: \url{https://github.com/avp-neelam/GraphLabelAlignment}

\begin{wrapfigure}{r}{2.5in}
  \vspace{-.2in}
  \centering
  \includegraphics[width=\linewidth]{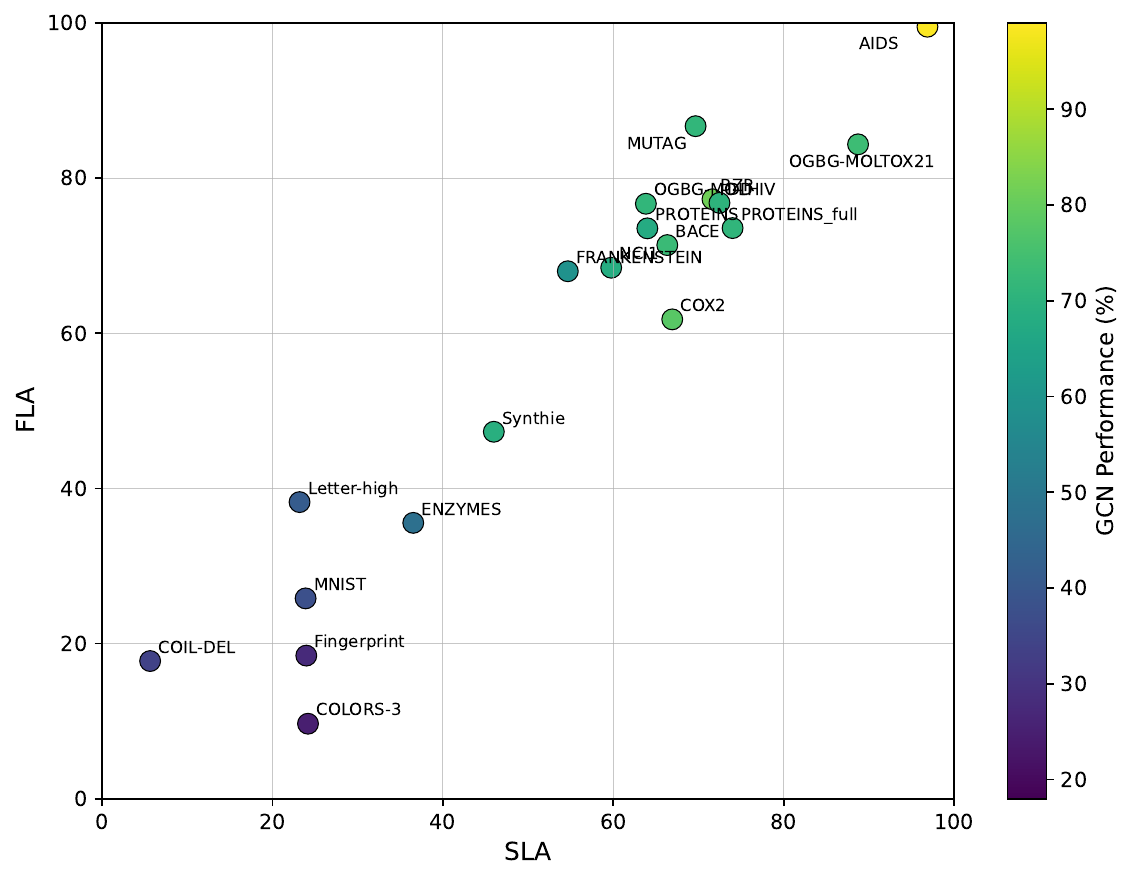}
  \caption{\footnotesize \textbf{Real benchmarks in the alignment plane (GCN).} Each point is a real graph classification dataset positioned by its SLA and FLA scores, colored by GCN accuracy. Other backbones are shown in Appendix~\Cref{fig:real-data-plane-all}.}
  \label{fig:gcn-plot}
  \vspace{-.3in}
\end{wrapfigure}

\paragraph{Real benchmark suite (1a).} We use a diverse set of graph classification datasets spanning molecules, proteins, social networks, and image-to-graph constructions (TU-style and OGB molecular graphs). Some datasets have node features (SLA and FLA); others are featureless (FLA undefined), in which case we report only SLA and omit AlignAvg. Dataset lists, statistics, and feature availability are in Appendix~\Cref{tab:dataset_statistics}.

\paragraph{Synthetic alignment sweep (1b).} To control alignment, we generate synthetic datasets by sweeping structural and feature separation $(s,f)$ as in \Cref{sec:synthetic}, covering structure-only, feature-only, and jointly-aligned regimes. We evaluate the same architectures and tuning protocol as on real benchmarks and compute SLA/FLA using the restricted-view probes.

\paragraph{Heterophilic ego-networks (2a).} To model interaction-driven classification, we convert heterophilic node benchmarks into ego-graph tasks: for each node, we extract the induced $k$-hop ego-graph and use the center-node label as the graph label. We include a center-node indicator feature in all ego-graph experiments. Construction and preprocessing are in Appendix~\Cref{app:ego_details}.

\paragraph{Twin synthetic datasets (2b).} We use Twins (Section~\ref{sec:synthetic}) to remove structure-only and feature-only shortcuts while preserving the interaction-only label signal. All classes share the same SBM distribution and feature multiset; labels differ only by role-based feature assignment (Linear, Radial, PageRank, Eigen) versus random permutation. We evaluate binary  and three-class settings. Unless noted, each dataset has 1000 graphs with 100 nodes; details are in Appendix~\ref{app:synthetic}.

\paragraph{Models and baselines.} We evaluate pooled graph classifiers: \textbf{GCN}, \textbf{GraphSAGE} (\textbf{SAGE}), \textbf{GAT}, and \textbf{GPS}. Each model produces node embeddings and applies a permutation-invariant readout, mean pooling by default, to obtain graph-level embeddings for classification; readout variants and hyperparameters are in Appendix~\Cref{app:experiments}. For Twins, we also evaluate \textbf{GCN+DiffPool} to test whether hierarchical pooling mitigates the interaction-only failure mode.

\section{Results}\label{sec:results}
\subsection{Alignment plane on real benchmarks}\label{subsec:real_alignment_results}
We first test whether our restricted-view alignment scores organize the performance of standard \emph{pooled} graph classifiers on real graph classification benchmarks. For each dataset, we report baseline accuracies for \textbf{GCN}, \textbf{SAGE}, \textbf{GAT}, and \textbf{GPS} together with the corresponding alignment scores (\textbf{SLA} and \textbf{FLA} when node features are available; \textbf{SLA} only for featureless datasets). (Implementation details for backbones, pooled readout, and the hyperparameter grid are in Appendix~\Cref{app:experiments}. Full dataset statistics are in Appendix~\Cref{tab:dataset_statistics}.)

\begin{figure}[t]
  \centering
  \includegraphics[width=.45\columnwidth]{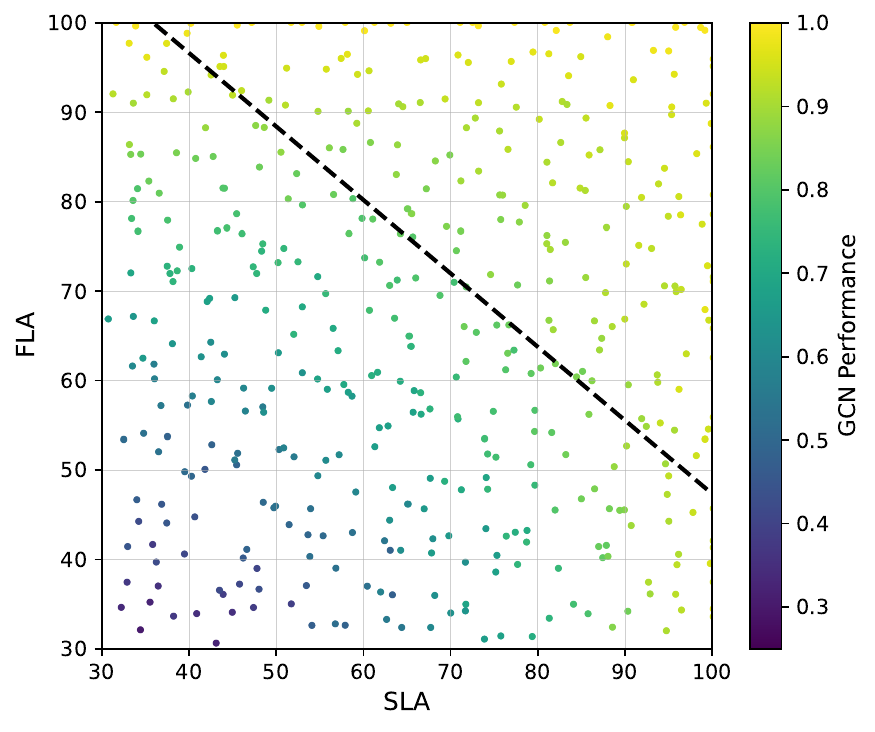}
    \includegraphics[width=.45\columnwidth]{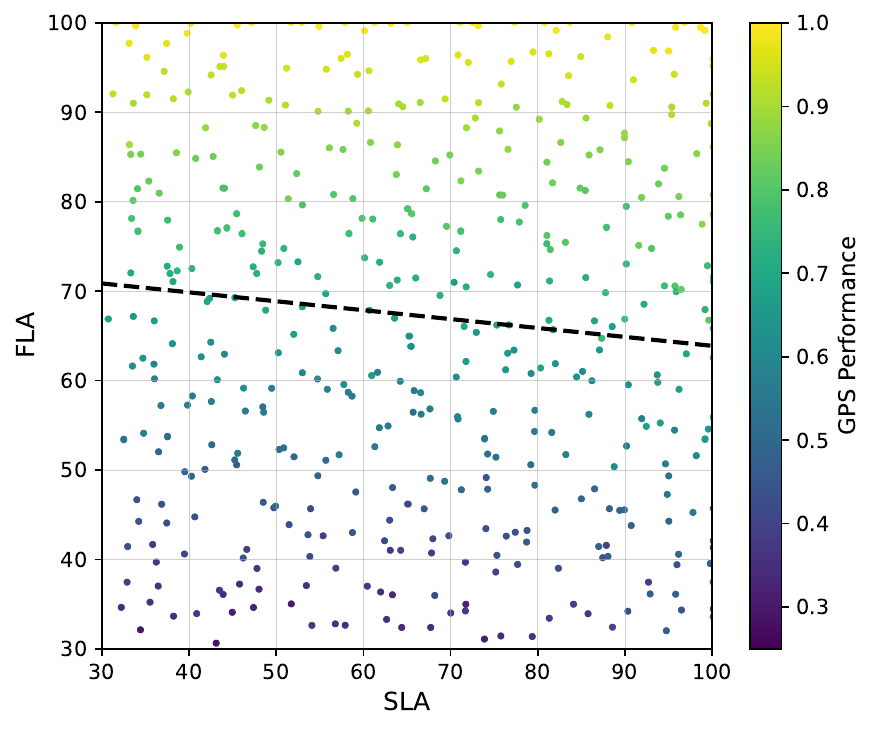}
  \caption{\textbf{Synthetic alignment sweep (GCN\& GPS).} Each point is a synthetic dataset positioned by its SLA and FLA, colored by GCN(left) and GPS (right) accuracy. The dashed line marks an illustrative 80\% accuracy contour. Other backbones are shown in Appendix~\Cref{fig:synth-plane}.}
  \label{fig:gcn-synthetic}
    \vspace{-.2in}
\end{figure}

\begin{wraptable}{r}{3in}
\vspace{-.2in}
\centering
\caption{\footnotesize Pearson correlations between alignment scores and model accuracy on 19 real benchmarks (datasets where FLA and AlignAvg ((SLA+FLA)/2) are defined; see \Cref{tab:gnn_baselines_extended}). Statistical significance: $^{*}p<0.05$, $^{**}p<0.01$, $^{***}p<0.001$.}
\label{tab:corr_all_alignment}
\footnotesize
\setlength{\tabcolsep}{5pt}
\begin{sc}
\resizebox{\linewidth}{!}{%
\begin{tabular}{l c c c c}
\toprule
\textbf{Alignment} & \textbf{GCN} & \textbf{SAGE} & \textbf{GAT} & \textbf{GPS} \\
\midrule
SLA      & 0.935$^{***}$ & 0.802$^{***}$ & 0.941$^{***}$ & 0.835$^{***}$ \\
FLA      & 0.928$^{***}$ & 0.715$^{***}$ & 0.961$^{***}$ & 0.759$^{***}$ \\
AlignAvg & 0.944$^{***}$ & 0.769$^{***}$ & 0.962$^{***}$ & 0.808$^{***}$ \\
\bottomrule
\end{tabular}}
\end{sc}
\vspace{-.15in}
\end{wraptable}

\paragraph{Key observation: a clear trend in the SLA--FLA plane.} Across featureful benchmarks, datasets with higher SLA and/or higher FLA tend to achieve higher pooled-model accuracy, while low-alignment datasets cluster in a region where all pooled architectures struggle. This trend is visually apparent in \Cref{fig:gcn-plot}, which places each dataset in the SLA--FLA plane and colors points by \textbf{GCN} accuracy. Importantly, the qualitative organization is not driven by a particular backbone: corresponding plots for \textbf{SAGE}, \textbf{GAT}, and \textbf{GPS} show the same large-scale pattern (Appendix~\Cref{fig:real-data-plane-all}). Throughout, we treat this as a descriptive organization of benchmark difficulty, rather than a causal claim about model behavior.

\paragraph{Quantifying the trend via correlations.} To quantify how well alignment predicts pooled accuracy beyond the qualitative plane plots, we compute Pearson correlations between each alignment score and model accuracy on the subset of real benchmarks where \textbf{FLA} (and thus \textbf{AlignAvg}) is defined.

\begin{table*}[t]
\centering
\caption{\footnotesize {\bf Real Graph Benchmarks.} Pooled-model accuracy (\%) tracked against the alignment plane: structural and feature alignment scores (SLA/FLA, \%). AlignAvg summarizes alignment as $(\mathrm{SLA}+\mathrm{FLA})/2$. Datasets are ordered by decreasing AlignAvg when FLA is defined, and by decreasing SLA for featureless datasets (FLA/AlignAvg shown as --).}
\label{tab:gnn_baselines_extended}
\footnotesize
\setlength{\tabcolsep}{5pt}
\begin{sc}
\resizebox{\textwidth}{!}{%
\begin{tabular}{l|cccc|ccc}
\toprule
\textbf{Dataset} & \textbf{GCN} & \textbf{SAGE} & \textbf{GAT} & \textbf{GPS} & \textbf{SLA} & \textbf{FLA} & \textbf{AlignAvg}$\downarrow$ \\
\midrule
AIDS            & 98.40\std{0.30} & 98.30\std{0.51} & 95.95\std{1.04} & 98.85\std{0.30} & 99.51 & 96.85 & 98.18 \\
MOLTOX21        & 73.45\std{0.41} & 73.63\std{0.53} & 71.95\std{0.40} & 73.70\std{1.01} & 84.35 & 88.71 & 86.53 \\
MUTAG           & 71.28\std{5.48} & 69.74\std{4.10} & 67.18\std{5.48} & 70.77\std{7.36} & 86.69 & 69.65 & 78.17 \\
DD              & 70.30\std{2.06} & 70.55\std{3.08} & 69.03\std{3.36} & 71.90\std{2.65} & 76.81 & 72.45 & 74.63 \\
BZR             & 81.48\std{3.58} & 80.00\std{3.61} & 83.46\std{4.66} & 83.21\std{4.39} & 77.28 & 71.63 & 74.46 \\
PROTEINS\_full  & 71.07\std{3.16} & 71.70\std{3.73} & 73.84\std{1.64} & 74.11\std{2.06} & 73.56 & 73.99 & 73.78 \\
MOLHIV          & 71.41\std{1.49} & 74.47\std{1.95} & 73.39\std{1.24} & 71.26\std{1.60} & 76.69 & 63.82 & 70.26 \\
BACE            & 72.76\std{1.87} & 77.17\std{0.53} & 60.53\std{2.25} & 69.74\std{2.21} & 71.37 & 66.33 & 68.85 \\
PROTEINS        & 67.59\std{3.50} & 68.12\std{2.22} & 67.05\std{4.72} & 68.93\std{3.62} & 73.52 & 63.99 & 68.76 \\
COX2            & 78.09\std{5.37} & 78.30\std{4.44} & 76.81\std{5.23} & 77.87\std{5.27} & 61.79 & 66.91 & 64.35 \\
NCI1            & 68.30\std{1.59} & 69.39\std{1.20} & 64.53\std{1.81} & 69.39\std{0.88} & 68.45 & 59.75 & 64.10 \\
FRANKENSTEIN    & 59.12\std{1.29} & 62.07\std{1.52} & 61.57\std{1.68} & 60.53\std{1.47} & 67.99 & 54.66 & 61.33 \\
Synthie         & 69.00\std{5.83} & 77.00\std{4.00} & 45.50\std{4.91} & 45.00\std{4.47} & 47.30 & 45.98 & 46.64 \\
ENZYMES         & 47.50\std{3.03} & 50.33\std{5.31} & 43.00\std{4.49} & 50.67\std{1.43} & 35.56 & 36.53 & 36.05 \\
Letter-High     & 41.38\std{1.70} & 79.29\std{1.80} & 30.80\std{2.47} & 81.51\std{1.36} & 38.23 & 23.19 & 30.71 \\
MNIST           & 37.85\std{0.70} & 48.80\std{0.61} & 32.01\std{0.76} & 47.33\std{0.62} & 25.83 & 23.90 & 24.87 \\
CIFAR-10        & 27.77\std{0.59} & 39.85\std{1.06} & 31.70\std{0.39} & 41.54\std{0.72} & 18.45 & 23.98 & 21.22 \\
COLORS-3        & 24.92\std{1.34} & 26.08\std{1.09} & 25.08\std{0.78} & 25.03\std{0.78} & 9.67  & 24.18 & 16.93 \\
COIL-DEL        & 33.97\std{0.92} & 55.59\std{0.59} & 18.85\std{1.42} & 46.08\std{2.96} & 17.75 & 5.66  & 11.71 \\
\midrule
REDDIT-B        & 72.25\std{2.98} & 50.15\std{1.14} & 50.65\std{0.94} & 50.15\std{1.14} & 89.01 & -- & -- \\
COLLAB          & 81.00\std{0.06} & 79.10\std{0.05} & 69.30\std{0.03} & 51.96\std{1.11} & 73.61 & -- & -- \\
IMDB-B          & 63.30\std{0.04} & 58.90\std{0.06} & 54.60\std{0.07} & 50.00\std{2.30} & 70.43 & -- & -- \\
REDDIT-5k       & 42.50\std{0.02} & 46.60\std{0.02} & 48.70\std{0.01} & 18.90\std{1.30} & 55.37 & -- & -- \\
IMDB-M          & 40.20\std{0.05} & 42.00\std{0.05} & 37.70\std{0.05} & 31.20\std{3.77} & 46.00 & -- & -- \\
\bottomrule
\end{tabular}}
\end{sc}
\vspace{-.15in}
\end{table*}

All four backbones show strong positive correlations with both \textbf{SLA} and \textbf{FLA}: labels that are more accessible from either restricted view tend to yield better pooled-model performance. Moreover, $\mathrm{AlignAvg}=(\mathrm{SLA}+\mathrm{FLA})/2$ is consistently among the most correlated summaries, providing a compact ordering for reporting and visualization. Since alignment scores are computed without training any GNN, these correlations serve as an architecture-agnostic diagnostic, suggesting that marginal label accessibility explains a substantial part of benchmark difficulty for these pooled models. Exact summary constructions, probe protocols, and complexity details are given in Appendix~\ref{app:summary_vectors}.

\paragraph{Robustness of the alignment plane}. Appendix~\ref{app:probe-validation} tests the design choices behind the scores. Removing structural blocks keeps SLA rank correlation with the full descriptor at 0.968 or above, and a 12-dimensional subset without persistent homology reaches 0.988. Richer feature summaries raise FLA on some low-alignment datasets, but keep rank correlation with centroid FLA at 0.934 or above. Across summaries and probes, no dataset moves directly between the high/high and low/low regimes, and correlations with accuracy keep their pattern across all four backbones. A stronger five-model probe suite flags \textsc{Letter-High} as the main probe-sensitive case. Appendix~\ref{app:screening} reports compute cost, leave-one-dataset-out prediction of GCN accuracy (Pearson $r=0.917$), a benchmark triage score, and scaffold splits, under which our correlations should be read as an in-distribution diagnostic.

\subsection{Synthetic sweep validates the alignment plane}\label{subsec:synth_sweep_results}
Real datasets vary simultaneously in many confounding factors (domain, size, feature type, and class structure), so we next test the alignment story in a controlled environment. We generate a grid of synthetic datasets by sweeping structural and feature separation parameters as described in \Cref{sec:synthetic}, thereby spanning structure-only, feature-only, and jointly-aligned regimes while holding other factors fixed. (Full generation details and how SLA/FLA track the control parameters are provided in Appendix~\ref{app:synthetic}, with full sweep tables reported at the end of the appendix.)

\paragraph{Key observation: accuracy contours follow the alignment geometry.} \Cref{fig:gcn-synthetic} visualizes the synthetic sweep in the alignment plane for \textbf{GCN}. The dashed curve marks an \emph{illustrative} boundary for ``good'' performance (80\% accuracy): datasets above this contour concentrate in high-alignment regions of the plane, while low-alignment regions remain challenging. As in the real benchmarks, the same qualitative organization is observed for other backbones (Appendix~\Cref{fig:synth-plane}), indicating that the alignment plane captures a stable performance geometry rather than an artifact of a single model.

\subsection{Twin datasets expose a pooling limitation}\label{subsec:interaction_results}
The preceding results show that pooled performance is largely organized by \emph{marginal} label accessibility. We now ask whether low alignment can also reflect a genuine interaction-dependent signal rather than noise. Twins answer this in a controlled setting: by construction, marginal structural and feature shortcuts are removed, while role--feature coupling remains discriminative.

\paragraph{Synthetic Twins isolate an interaction-only failure mode.} We study \emph{Twins} (\Cref{sec:synthetic}), where all classes share the same SBM structure distribution and the same multiset of node feature types. Balanced binary and three-class variants differ only in how feature types are assigned to nodes according to a role score (Linear, Radial, PageRank, Eigen) versus a random permutation. Thus structure-only and feature-only probes are non-informative by design, and any gain over chance requires preserving global role--feature correspondence. Construction details, role-score definitions, and alignment suppression checks are given in Appendix~\Cref{app:twins_details,app:twins_roles,app:twins_alignment,app:twins_classes}.

The result is clear: across all role-score variants and both binary and three-class settings, \textbf{pooled GNNs and GPS remain near random guessing} ($50\%$ binary, $33\%$ three-class), with the failure most pronounced in the three-class setting. To test whether hierarchical pooling mitigates this effect, we also evaluate \textbf{GCN+DiffPool}; it remains near chance across Twins variants. These results suggest a correspondence bottleneck in standard pooled graph-classification pipelines: progress in this regime likely requires readouts or objectives that preserve the relation between node roles and feature assignments.

\paragraph{Heterophilic ego-graphs as realistic interaction-driven instances.} We convert heterophilic node-classification graphs into ego-graph classification tasks (Appendix~\ref{app:ego_details}) as a noisier but more realistic corroboration. These datasets are \emph{interaction-dominated}: the class signal can depend on the arrangement of features around the center node rather than on global structure or pooled feature composition alone. Results are summarized in \Cref{tab:webkb_actor_khop} and analyzed in Appendix~\ref{app:ego_results}. Pooled message-passing baselines achieve low accuracy throughout, typically $\sim$25--45\%, consistent with the Twins finding. GPS is a consistent outlier on WebKB at small $k$, achieving up to 60\% on Cornell at $k=1$ despite $\mathrm{AlignAvg}\approx 25$. We discuss this boundary case in Appendix~D.1, where global attention and center-node information may capture signal not reflected by marginal alignment scores.

Together, Twins and ego-graphs show that low alignment need not mean absence of signal: it can mark regimes where the discriminative information lies in structure--feature correspondence that standard pooled pipelines do not reliably preserve.

\begin{table*}[t]
\centering
\caption{\textbf{Twin setting} Accuracy results (\%) for binary (random assignment vs.\ one structured assignment) and  triclass tasks (random assignment plus two structured assignments).}
\label{tab:twin_results}
\begin{sc}
\footnotesize
\setlength{\tabcolsep}{4pt}
\resizebox{\textwidth}{!}{%
\begin{tabular}{lcccc|cccccc}
 & \multicolumn{4}{c}{\normalsize \textbf{Binary}} & \multicolumn{6}{c}{\normalsize \textbf{Triclass}} \\
\cmidrule(lr){2-5} \cmidrule(lr){6-11} 

& \textbf{Linear} & \textbf{Radial} & \textbf{PageRank} & \textbf{Eigen}
& \textbf{Lin \& Rad} & \textbf{Lin \& PR} & \textbf{Lin \& Eigen}
& \textbf{Rad \& PR} & \textbf{Rad \& Eigen} & \textbf{PR \& Eigen} \\
\midrule
FLA  & 24.73 & 24.73 & 24.73 & 24.73 & 12.02 & 12.02 & 12.02 & 12.02 & 12.02 & 12.02 \\
SLA  & 27.27 & 27.27 & 27.27 & 27.27 & 14.11 & 14.11 & 14.11 & 14.11 & 14.11 & 14.11 \\
\midrule
GCN  & 46.92 & 46.92 & 46.17 & 49.83 & 30.67 & 30.44 & 31.06 & 32.44 & 31.33 & 32.00 \\
SAGE & 49.50 & 48.50 & 48.67 & 49.50 & 33.11 & 32.39 & 31.56 & 32.00 & 33.00 & 33.11 \\
GAT  & 47.92 & 45.58 & 44.00 & 46.83 & 31.78 & 28.56 & 31.00 & 31.06 & 31.39 & 31.33 \\
GPS  & 48.50 & 49.67 & 49.25 & 49.33 & 32.06 & 32.50 & 32.94 & 32.67 & 32.83 & 32.22 \\
DiffPool & 43.00 & 45.25 & 45.75 & 46.25 & 27.78 & 30.22 & 27.56 & 29.44 & 28.72 & 28.94 \\
\bottomrule
\end{tabular}}
\end{sc}
\vspace{-.15in}
\end{table*}

\subsection{Discussion and takeaways}\label{sec:takeaways}
Overall, our experiments support the following takeaways.

\paragraph{Alignment is a practical organizer and diagnostic of pooled-model learnability.} On real benchmarks, pooled-model accuracy is strongly structured by the SLA--FLA plane, and the same geometry appears under controlled synthetic sweeps where structural and feature signal can be dialed independently. This makes alignment a simple, architecture-agnostic way to diagnose where label signal is accessible from restricted views.

\paragraph{Low alignment is not fixed by swapping encoders.} When both marginal views are weak, all pooled architectures degrade substantially and can approach near-chance behavior, suggesting that encoder choice alone is insufficient in interaction-dominated regimes; the bottleneck can lie in what standard pooled pipelines preserve.

\paragraph{Interaction-only tasks reveal a correspondence bottleneck.} Heterophilic ego-graphs and Twins instantiate regimes where labels depend on how features are arranged relative to structure. In Twins, where marginal shortcuts are eliminated by design, pooled models remain near chance, indicating that permutation-invariant readout can discard global coupling signatures. This motivates alignment-aware evaluation and, in interaction-dominated settings, objectives or readouts that better preserve role-level structure--feature coupling beyond standard pooling. (For an optional backbone-aware alignment predictor and a practical decision map tying SLA/FLA regimes to backbone choice, see Appendix~\ref{app:gla-recipe}.)

\section{Conclusion}\label{sec:conclusion}
This work takes a data-centric view of graph classification by studying where label signal is accessible before training a graph encoder. Through \emph{graph label alignment}, we show a consistent pattern across real benchmarks, heterophilic ego-graphs, and controlled synthetic families: when labels are accessible from structure or pooled features, standard pooled pipelines tend to perform well, whereas low alignment marks regimes where changing the encoder alone yields limited gains. Our interaction-focused stress tests further reveal a correspondence bottleneck in standard pooled pipelines, where global feature-role coupling can be lost despite expressive message passing. More broadly, these results suggest that graph benchmark analysis should distinguish marginal shortcuts from interaction-dependent signal. Future directions include readouts and objectives that preserve structure--feature correspondence, alignment-aware benchmark design, and extensions of this diagnostic view to link prediction, retrieval, and transfer.

\begin{ack}
This work was partially supported by National Science Foundation under grant DMS-2229417, OAC-2115094,  CNS-2331424, ITE-2452833, Simons Foundation under grant SFM-21269, ARL/Army Research Office awards W911NF-24-1-0202 and W911NF-24-2-0114, and Virginia Commonwealth Cyber Initiative grants. The authors acknowledge the Texas Advanced Computing Center (TACC) at  UT Austin for providing computational resources that have contributed to the research results reported within this paper. 
\end{ack}

\bibliographystyle{alpha}
\bibliography{references}
\normalsize

\clearpage
\appendix
\section*{Appendix}
\noindent\textbf{Appendix roadmap.}

\Cref{app:experiments} gives \textbf{reproducibility} details, plus dataset metadata and per-backbone SLA--FLA plots.

\Cref{app:summary_vectors} defines \textbf{SLA/FLA} summaries, probes, and complexity.

\Cref{app:probe-validation} reports \textbf{structural and feature ablations}, a \textbf{probe sufficiency self-check}, and \textbf{regime and seed stability}.

\Cref{app:ego_details} covers \textbf{ego-graphs} (construction and extra results).

\Cref{app:synthetic} details \textbf{synthetics} (sweep, coverage, Twins).

\Cref{app:background} reviews \textbf{signatures} (topology and spectrum).

\Cref{app:gla-recipe} reports \textbf{GLA} (predictor, decision map, backbone rules), with full sweep tables at the end.

\Cref{app:screening} reports \textbf{screening use} (compute cost, out-of-sample prediction, $\Delta_{\mathrm{graph}}$) and \textbf{scaffold splits}.

\section{Further Experimental Details} \label{app:experiments}
\paragraph{Implementation Details.} We evaluated performance across two families of GNNs: (i) message-passing GNNs (GCN, GraphSAGE, and GAT); and (ii) a hybrid local-global GPS architecture. All models operate on node features, producing predictions via global pooling.

\paragraph{Message-passing GNNs.} GCN~\cite{kipf2016semi}, GraphSAGE~\cite{hamilton2017inductive}, and GAT~\cite{velivckovic2017graph} use a unified $l$-layer message-passing architecture with hidden dimension $h$. GCN and SAGE layers map using mean aggregation. For GAT, hidden layers use multi-head attention with $H$ heads, yielding intermediate representation with dimension $h$. Each convolution is followed by batch normalization, a nonlinearity (ReLU for GCN/SAGE, ELU for GAT), and dropout with rate $p$. Node embeddings are aggregated using global mean pooling and passed through a linear classifier after the final layer.

\paragraph{GPS.} GPS architecture~\cite{rampavsek2022recipe} combines local message passing and global attention within each layer. Input features are first projected to dimension $h$ via a linear map. Each of the $l$ layers consist of a GraphSAGE convolution as the local MPNN component together with multi-head global attention with $H$ heads. Outputs are normalized, passed through ReLU nonlinearity, and dropout with rate $p$. Residual connections are applied at each layer, updating node states. The final embeddings are pooled at the graph level and mapped to class logits using a linear classifier. 

\paragraph{Grid Search for Hyperparameters.} For real and heterophilic ego-graph datasets we run a grid search over hyperparameters: $l=2,3$ layers; $h=32, 64, 128$ hidden dimension; $H=1,2,4,8$ multi-head attention (when relevant); $p=0.3,0.5$ dropout; and learning rate of $1\times10^{-3}$ with early stopping on validation 

\begin{table*}[t]
    \centering
    \caption{Dataset statistics across molecular, computer vision, synthetic, and social graph benchmarks.}
    \label{tab:dataset_statistics}
    \footnotesize
    \begin{sc}
    \resizebox{\textwidth}{!}{%
    \begin{tabular}{llccccc}
    \toprule
    \textbf{Group} & \textbf{Dataset} & \textbf{\#Graphs} & \textbf{Av.Nodes} & \textbf{Av.Edges} & \textbf{\# Feat.} & \textbf{\#Classes} \\
    \midrule
    \multirow{13}{*}{Molecular} 
    & AIDS            & 2000  & 15.69  & 32.39   & 42  & 2 \\
    & BACE            & 1513  & 34.09  & 73.72   & 9   & 2 \\
    & BZR             & 405   & 35.75  & 76.72   & 56  & 2 \\
    & COX2            & 467   & 41.22  & 86.89   & 38  & 2 \\
    & DD              & 1178  & 284.32 & 1431.32 & 89  & 2 \\
    & ENZYMES         & 600   & 32.60  & 124.27  & 21  & 6 \\
    & FRANKENSTEIN    & 4337  & 16.90  & 35.75   & 780 & 2 \\
    & MOLHIV     & 41127 & 25.51  & 54.94   & 9   & 2 \\
    & MOLTOX21   & 7831  & 18.57  & 38.59   & 9   & 2 \\
    & MUTAG           & 188   & 17.93  & 39.59   & 7   & 2 \\
    & NCI1            & 4110  & 29.87  & 64.60   & 37  & 2 \\
    & PROTEINS        & 1113  & 39.06  & 145.63  & 4   & 2 \\
    & PROTEINS\_full  & 1113  & 39.06  & 145.63  & 32  & 2 \\
    \midrule
    \multirow{4}{*}{Vision}
    & CIFAR-10        & 45000 & 117.65 & 941.18  & 3   & 10 \\
    & COIL-DEL        & 3900  & 21.54  & 108.47  & 2   & 100 \\
    & Letter-high     & 2250  & 4.67   & 9.00    & 2   & 15 \\
    & MNIST           & 55000 & 70.56  & 564.50  & 1   & 10 \\

    \midrule
    \multirow{2}{*}{Synthetic}
    & COLORS-3        & 10500 & 61.31  & 182.05  & 5   & 11 \\
    & Synthie         & 400   & 95.00  & 345.86  & 15  & 4 \\
    \midrule
    \multirow{5}{*}{Social}
    & COLLAB          & 5000  & 74.49  & 4914.43 & 0   & 3 \\
    & IMDB-B          & 1000  & 19.77  & 193.06  & 0   & 2 \\
    & IMDB-M          & 1500  & 13.00  & 131.87  & 0   & 3 \\
    & REDDIT-BINARY   & 2000  & 429.63 & 995.51  & 0   & 2 \\
    & REDDIT-5K       & 4999  & 508.52 & 1189.75 & 0   & 5 \\
    \bottomrule
\end{tabular}}
\end{sc}
\end{table*}


\begin{figure*}[t]
  \begin{center}
    \centerline{
        \includegraphics[width=0.5\columnwidth]{figures/plots/gcn.pdf} \includegraphics[width=0.5\columnwidth]{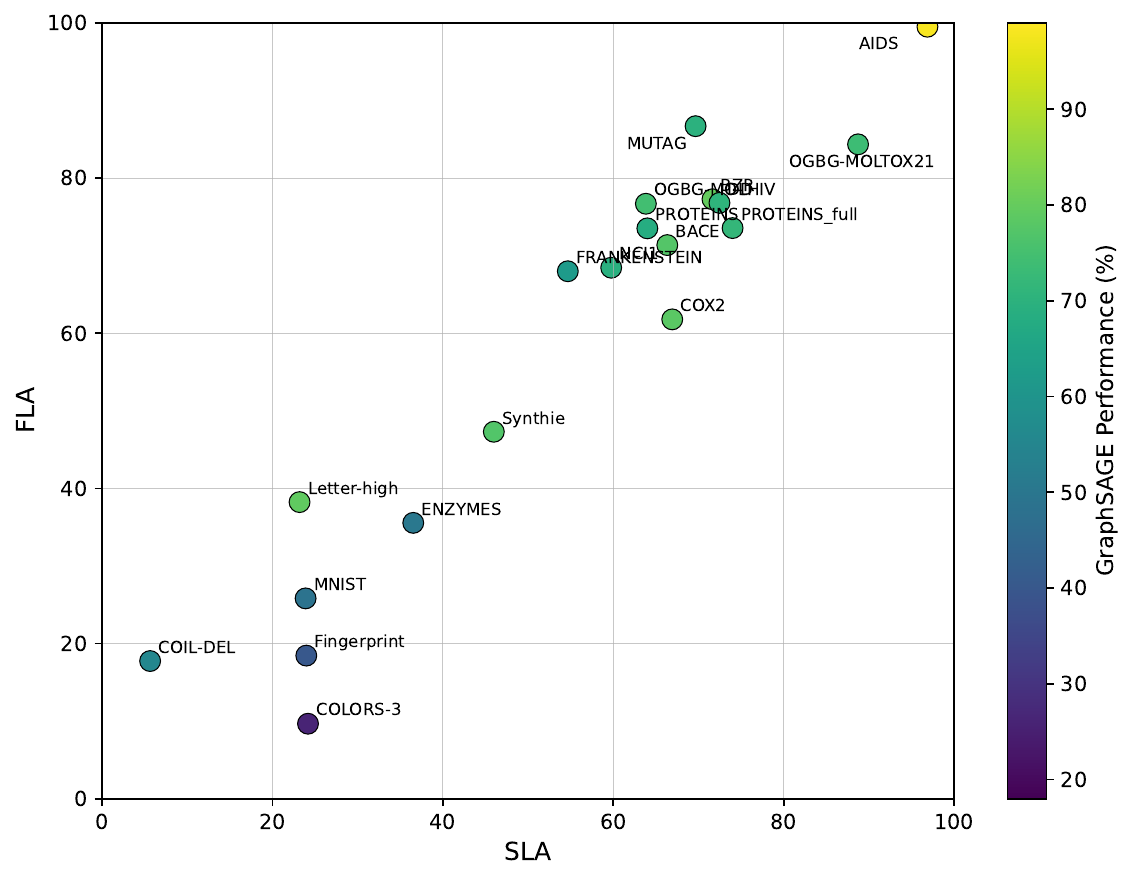}
    }\centerline{
        \includegraphics[width=0.5\columnwidth]{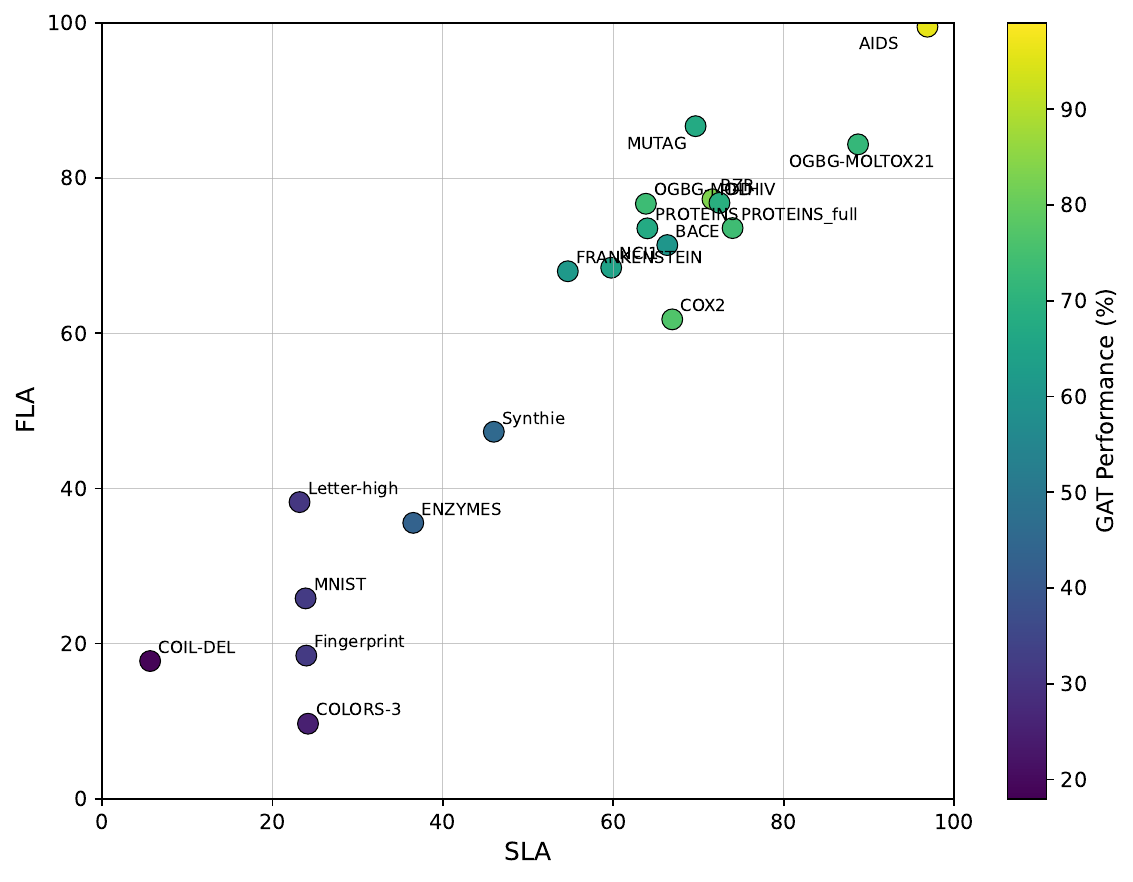} \includegraphics[width=0.5\columnwidth]{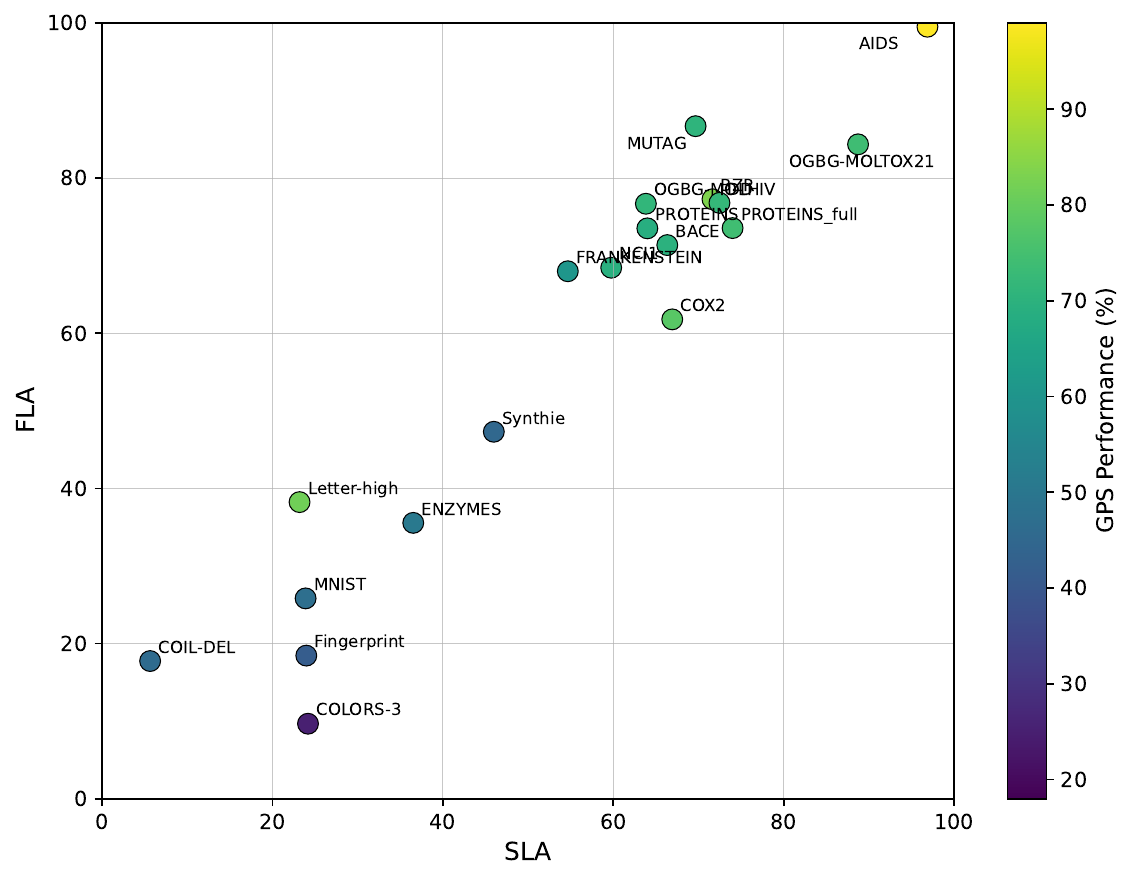}
    }
    \caption{\textbf{GNN on graph classification benchmarks.} Each point is a real benchmark dataset in the SLA-FLA plane, colored by GNN classification accuracy. Beginning in the top left: GCN, GraphSAGE, GAT, and GPS.}
    \label{fig:real-data-plane-all}
  \end{center}
\end{figure*}

\section{Structural and Feature Label Alignment}\label{app:summary_vectors}
This appendix specifies the exact construction and implementation details of the \emph{Structural Summary Vectors} and \emph{Feature Summary Vectors} used to compute \textbf{Structure Label Alignment (SLA)} and \textbf{Feature Label Alignment (FLA)}, along with their computational complexity.

Let $\mathcal{D} = \{(G_i, X_i, y_i)\}_{i=1}^N$ denote a graph classification dataset, where each graph $G_i = (V_i, E_i)$ has node features $X_i \in \mathbb{R}^{|V_i| \times d}$ and class label $y_i$.

\subsection{Structural Summary Vectors}\label{app:struct_summary}
For each graph $G = (V,E)$, we construct a fixed-dimensional \emph{Structural Summary Vector} $\phi_{\text{struct}}(G) \in \mathbb{R}^{48}$ by concatenating topological, spectral, and global structural statistics. Formally:
\[\phi_{\text{struct}}(G) = \left[\beta_{0,1}(G) \,|\,|\; \lambda(G) \,|\,|\; \mu(G) \,|\,|\; g(G) \right]\]
where $\beta_{0,1}(G)$ acts as the topological summary; $\lambda(G)$ and $\mu(G)$ the spectral summary; and $g(G)$ contains global structural statistics.

\paragraph{Topological Summary.} For each graph we utilize sublevel-filtration based PH with degree as the filtration function, on ten quantiled bins, and compute the Betti-0 and Betti-1 numbers. To improve computational complexity we compute $\beta_0$ as the number of connected components and $\beta_1$ by Euler characteristic. The resulting concatenation
\[\beta_{0,1}(G) = \left(\beta_0(G^1),\beta_0(G^2),\ldots,\beta_0(G^{10}),\beta_1(G^1),\beta_1(G^2),\ldots,\beta_1(G^{10}),\right) \in \mathbb{R}^{20}\]
is the defined topological summary included in $\phi_{\text{struct}}(G)$.

\paragraph{Spectral Summary.} Let $A$ be the adjacency matrix of $G$ and $L=I-D^{-1/2}AD^{-1/2}$ the normalized Laplacian. To compute $\lambda(G)\in\mathbb{R}^4$ we find the four smallest (nonzero) eigenvalues of $L$, similarly $\mu(G)\in\mathbb{R}^4$ are the four largest eigenvalues of $A$. To maintain fixed length summary vectors across graphs, if a graph has fewer than four eigenvalues all available values are used and the remaining length is padded with zeros~\cite{chung1997spectral}.

\paragraph{Global Summary.} The global summary vector $g(G)\in\mathbb{R}^{20}$ contains number of nodes, number of edges, graph density, and degree quantiles at $5\%$, $25\%$, $50\%$, $75\%$, and $95\%$. Additionally, we report mean, standard deviation, skewness, and excess kurtosis for node-level degree, clustering coefficient, and triangle count.

\subsection{Feature Summary Vectors}\label{app:feat_summary}
For each graph $G$ with present node feature matrix $X\in\mathbb{R}^{|V|\times d}$, we construct a \emph{Feature Summary Vector} $\phi_{\text{feat}}(G)\in\mathbb{R}^d$. Our reference summary is the centroid of the corresponding feature-space point-cloud. That is,
\[\phi_{\text{feat}}(G) = \frac{1}{|V|} \sum_{v\in V} X_v.\]

We use the centroid because it is the cheapest summary and its FLA relates to GNN accuracy about as strongly as the richer summaries do, not because it scores highest. Richer permutation-invariant summaries often raise FLA itself; \Cref{app:fla-ablation} reports this comparison.

\subsection{Alignment Score Computation and Model Choice}\label{app:align-score}
We detail the construction and evaluation protocol for the structural and feature alignment scores introduced in \cref{sec:method}. From the fixed-length summary vectors $\phi_{\text{struct}}$ and $\phi_{\text{feat}}$, we obtain two dataset-level design matrices $X_S$ and $X_F$, with class labels $y$ inherited from the original graph classification task.

\paragraph{Baseline probe suite.} In the main experiments, we consider a fixed suite of two elementary classifiers: logistic regression and an RBF-kernel SVM\footnote{After early ablations, we selected these two models to probe both linear separability (LogReg) and simple non-linear separability (RBF-SVM~\cite{cortes1995support}) of the summary vectors, while keeping the probes architecture-agnostic and shallow.}. Scores are computed using 5 random seeds and 10-fold stratified cross-validation; we report mean accuracy across seeds and folds to remain consistent with the evaluation of downstream GNN baselines. The final SLA and FLA values are defined as the average performance of the LogReg and RBF-SVM probes on the corresponding structural or feature summaries.

\paragraph{Limitations of the current probe models.} While this probe suite provides a simple and interpretable baseline, it has several practical limitations. First, the RBF-SVM does not scale favorably with the number of graphs $N$ or with increasing summary density, requiring $\mathcal{O}(N^2)$ memory for the kernel matrix and up to $\mathcal{O}(N^3)$ time in the worst case. As a result, SLA/FLA evaluation can become a computational bottleneck for large graph collections or dense synthetic sweeps. Second, although logistic regression is computationally efficient, it probes only linear decision boundaries and may underestimate alignment when the label signal is weakly non-linear in the summary space. Conversely, the RBF-SVM can overemphasize fine-grained separability that may not translate to realistic graph models, particularly when the summaries are low-dimensional but densely clustered.

\paragraph{Alternative probe suites and alignment planes.} Importantly, SLA and FLA are not tied to a specific choice of classifier. The alignment framework can be instantiated with alternative probe suites, thereby inducing different \emph{alignment planes}. For example, replacing the RBF-SVM with cheaper non-linear probes (e.g., random-feature approximations, linear models on polynomial expansions, or small shallow MLPs) yields a computationally lighter alignment evaluation that scales linearly in $N$. Such variants trade expressivity for efficiency and can be preferable when screening large numbers of datasets or synthetic configurations. Throughout this work, we treat the $(\widehat{\mathrm{SLA}}, \widehat{\mathrm{FLA}})$ plane defined by LogReg+RBF-SVM as a reference instantiation, but emphasize that the framework is modular and admits alternative probe choices depending on computational budget and desired notion of separability.

\subsection{Computational Complexity and Classifier Selection}\label{app:gla-complexity}
We analyze the computational complexity of the SLA/FLA evaluation pipeline used throughout this work. The procedure consists of two main stages: (i) graph-level summary construction and (ii) repeated classifier evaluation under $s=5$ random seeds using stratified cross-validation.

\paragraph{Notation.} Let $N$ denote the number of graphs in a dataset. For graph $i$, let $n_i = |V_i|$ and $m_i = |E_i|$ be the number of nodes and edges, respectively. Let $F$ denote the node feature dimension. The number of graph classes is denoted by $C$. Structural summaries are constructed using $T=10$ filtration thresholds and $k=4$ eigenvalues.

\paragraph{Structural Summaries.} For each graph, the structural summary includes Betti profiles, spectral features, and standard graph statistics:
\begin{itemize}
    \item \textbf{Betti profiles.} Using a degree-based filtration with $T=10$ thresholds, each threshold requires scanning all edges and computing connected components, yielding a total cost of
    \[\Theta\!\left(T (n_i + m_i)\right).\]
    \item \textbf{Spectral features.} The top $k=4$ eigenvalues of both the normalized Laplacian and the adjacency matrix are computed using sparse Lanczos methods. Each eigensolve has cost
    \[\tilde{\Theta}\!\left(k (n_i + m_i)\right),\]
    where $\tilde{\Theta}(\cdot)$ suppresses iteration-dependent constants. Two such eigensolves are performed per graph.
    \item \textbf{Graph statistics.} Degree moments are linear in $n_i$, while clustering coefficients and triangle counts require
    \[\Theta\!\left(\sum_{v \in V_i} d(v)^2\right),\]
    which dominates in graphs with skewed degree distributions.
\end{itemize}

Combining these terms, the total cost of structural summary construction is
\[\sum_{i=1}^N \left[\Theta\!\left(T (n_i + m_i)\right)+ \tilde{\Theta}\!\left(k (n_i + m_i)\right)+ \Theta\!\left(\sum_{v \in V_i} d(v)^2\right)\right].\]
Since $T$ and $k$ are fixed constants, the resulting structural feature dimension is constant with respect to graph size.

\paragraph{Feature summaries.} For each graph, the feature summary is computed as the mean of node features, which costs $\Theta(n_i F)$. Summed over the dataset, the total cost is
\[\Theta\!\left(\sum_{i=1}^N n_i F\right),\]
and the resulting feature matrix has size $N \times F$.

\paragraph{Classifier Evaluation (SLA and FLA).} For both SLA and FLA representations, classifiers are evaluated using $s=5$ random seeds and $K=10$ stratified folds. Two models are used: logistic regression and an RBF-kernel SVM.

\emph{Logistic regression.} Training logistic regression using LBFGS requires
\[\Theta\!\left(I_{\mathrm{LR}} \, N d C\right)\]
per fold, where $d$ is the feature dimension and $I_{\mathrm{LR}}$ is the number of optimization iterations. Prediction is linear in $Nd$.

\emph{RBF-kernel SVM.} Training an RBF SVM requires between $\tilde{\Theta}(N^2)$ and $\Theta(N^3)$ time per fold, with $\Theta(N^2)$ memory in the worst case due to kernel matrix storage. This step becomes the dominant cost as $N$ grows.

\paragraph{Overall Complexity.} Let $d_s$ and $d_f$ denote the dimensions of the structural and feature summaries, respectively. The total time complexity of the SLA/FLA evaluation pipeline is
\[T_{\mathrm{total}} = T_{\mathrm{pre}} + 2 s K\left[\Theta(N d)+\Theta\!\left(I_{\mathrm{LR}} N d C\right)+T_{\mathrm{SVM}}(N)\right],\]
where $d \in \{d_s, d_f\}$ and $T_{\mathrm{SVM}}(N) \in [\tilde{\Theta}(N^2), \Theta(N^3)]$. The preprocessing cost $T_{\mathrm{pre}}$ is dominated by structural summary construction.

\paragraph{Memory Complexity.} Feature storage requires $\Theta(N d_s + N d_f)$ memory. During cross-validation, logistic regression requires $\Theta(dC)$ parameters, while the RBF SVM may require up to $\Theta(N^2)$ memory, which is the primary memory bottleneck.

\section{Probe Robustness and Structural Ablations}\label{app:probe-validation}
This appendix tests how much the alignment plane depends on our design choices: (i) which components of the structural summary vector contribute to SLA; (ii) whether a low SLA or FLA score may be limited by the reference probe suite; (iii) whether the centroid summary biases FLA; and (iv) whether regimes, correlations, and scores hold up when these choices and the random seeds change.

\paragraph{Benchmark sets.} The ablations in \Cref{app:struct-ablation,app:fla-ablation,app:regime-stability,app:seed-stability} and \Cref{tab:probe_summary} use 19 TU benchmarks: 14 with node attributes and 5 featureless social graphs. FLA and AlignAvg are defined only on the 14. This is not the same set as the 19 FLA-defined benchmarks in \Cref{tab:corr_all_alignment}, which also include the OGB molecular and vision datasets but not the social graphs, so correlations reported here are not directly comparable to that table. Re-computed full-descriptor SLA differs from \Cref{tab:gnn_baselines_extended} by at most 0.35 points.

\subsection{Structural Summary Ablations}\label{app:struct-ablation}
The structural summary vector $\phi_{\mathrm{struct}}(G) \in \mathbb{R}^{48}$ concatenates three families of permutation-invariant graph descriptors: a degree-filtration Betti profile (PH, 20 dimensions), a compact spectral signature (Spectral, 8 dimensions), and global graph statistics (Stats, 20 dimensions). To assess the contribution of each component, we evaluate SLA using each block in isolation and each pair of blocks, and compare against the full 48-dimensional vector.

\paragraph{Protocol.} Each ablation uses the same probe suite and cross-validation protocol as the main experiments (Appendix~\ref{app:align-score}): 5 seeds, 10-fold stratified cross-validation, and the average of logistic regression and RBF-kernel SVM. The only change across rows is the structural input representation used by the probe.

\begin{table}[t]
\centering
\caption{\footnotesize Structural ablation: SLA (\%) for each block alone, each pair, and the full 48-dimensional summary vector, on 14 TU benchmarks with node attributes and 5 featureless social benchmarks. \textbf{Bold} is the best score per row.}
\label{tab:struct_ablation}
\footnotesize
\setlength{\tabcolsep}{5pt}
\begin{sc}
\resizebox{\textwidth}{!}{%
\begin{tabular}{l|ccc|ccc|c}
\toprule
\textbf{Dataset} & \textbf{PH} (20d) & \textbf{Spectral} (8d) & \textbf{Stats} (20d) & \textbf{PH+Spec} (28d) & \textbf{PH+Stats} (40d) & \textbf{Spec+Stats} (28d) & \textbf{Full} (48d) \\
\midrule
AIDS & 99.39 & \textbf{99.62} & 99.50 & 99.59 & 99.44 & 99.55 & 99.53 \\
MUTAG & 86.70 & \textbf{87.61} & 86.50 & 86.38 & 86.39 & 86.13 & 86.49 \\
DD & 73.11 & 75.37 & 76.92 & 75.77 & 76.55 & \textbf{77.53} & 76.85 \\
BZR & 70.41 & 77.09 & 71.32 & 75.68 & 75.07 & 76.86 & \textbf{77.34} \\
PROTEINS\_full & 69.41 & 73.67 & 73.32 & 72.63 & 72.72 & \textbf{73.83} & 73.53 \\
PROTEINS & 69.41 & 73.66 & 73.32 & 72.57 & 72.72 & \textbf{73.83} & 73.48 \\
COX2 & 46.94 & \textbf{64.78} & 46.93 & 59.00 & 48.70 & 60.40 & 61.77 \\
NCI1 & 65.79 & 64.22 & 67.17 & 67.08 & 67.60 & 68.01 & \textbf{68.49} \\
FRANKENSTEIN & 66.10 & 62.73 & 67.26 & 66.64 & 67.38 & 67.99 & \textbf{68.04} \\
Synthie & 50.18 & 49.17 & 48.73 & \textbf{52.15} & 47.03 & 50.60 & 46.95 \\
ENZYMES & 28.18 & 25.25 & 34.00 & 31.72 & 33.95 & \textbf{35.70} & 35.68 \\
Letter-High & 31.55 & 36.13 & 38.40 & 38.40 & 37.97 & \textbf{38.76} & 38.29 \\
COLORS-3 & \textbf{9.83} & 9.45 & 9.81 & 9.37 & 9.72 & \textbf{9.83} & 9.60 \\
COIL-DEL & 10.02 & 16.05 & 16.05 & 14.35 & 16.14 & \textbf{18.42} & 17.76 \\
\midrule
REDDIT-B & 77.02 & 79.43 & 85.83 & 84.64 & 86.09 & 88.65 & \textbf{89.00} \\
COLLAB & 69.71 & 66.23 & 72.41 & 70.00 & 73.51 & 73.07 & \textbf{73.61} \\
IMDB-B & 66.23 & 69.58 & \textbf{71.23} & 70.19 & 70.70 & 71.04 & 70.55 \\
REDDIT-5k & 48.01 & 48.39 & 54.41 & 51.80 & 54.56 & 55.21 & \textbf{55.37} \\
IMDB-M & 42.70 & 42.73 & \textbf{46.70} & 45.71 & 45.85 & 46.61 & 45.83 \\
\midrule
\textbf{Average} & 56.88 & 59.01 & 59.99 & 60.19 & 60.11 & \textbf{61.69} & 61.48 \\
\bottomrule
\end{tabular}}
\end{sc}
\end{table}

\begin{table}[t]
\centering
\caption{\footnotesize Structural ablation summary. Rank correlations are Spearman against the full descriptor, for SLA (19 datasets) and for AlignAvg (the 14 datasets where FLA is defined, with centroid FLA held fixed). Pearson is against GCN accuracy on the 19 datasets. Regime agreement uses per-axis tertiles of the full-descriptor SLA.}
\label{tab:sla_summary}
\footnotesize
\setlength{\tabcolsep}{5pt}
\begin{sc}
\resizebox{\textwidth}{!}{%
\begin{tabular}{lcccccc}
\toprule
\textbf{Blocks} & \textbf{Dim} & \textbf{Avg SLA} & \textbf{SLA $\rho$} & \textbf{AlignAvg $\rho$} & \textbf{Pearson (GCN)} & \textbf{Regime agr.\ (\%)} \\
\midrule
PH & 20 & 56.88 & 0.987 & 0.982 & 0.868 & 73.7 \\
Spectral & 8 & 59.01 & 0.981 & 0.996 & 0.898 & 78.9 \\
Stats & 20 & 59.99 & 0.968 & 0.982 & 0.840 & 78.9 \\
PH+Spectral & 28 & 60.19 & 0.984 & 0.996 & 0.887 & 89.5 \\
PH+Stats & 40 & 60.11 & 0.994 & 0.982 & 0.850 & 84.2 \\
Spectral+Stats & 28 & 61.69 & 0.993 & 0.996 & 0.882 & 78.9 \\
Full & 48 & 61.48 & 1.000 & 1.000 & 0.880 & 100.0 \\
\bottomrule
\end{tabular}}
\end{sc}
\end{table}

\paragraph{Findings.} \Cref{tab:struct_ablation} shows that the best individual component varies by dataset and graph domain. Comparing the three single blocks and the full vector, Spectral is strongest on five datasets, predominantly molecular and biological graphs (AIDS, MUTAG, PROTEINS, PROTEINS-full, COX2). Stats is strongest on four (DD, Letter-High, IMDB-B, IMDB-M). PH leads on Synthie and COLORS-3, though on COLORS-3 all variants sit near chance ($\approx$9--10\% over 11 classes), so structure carries essentially no signal there. The full descriptor is best on the remaining eight. On COX2, Spectral (64.78) outperforms the full vector (61.77) by a notable margin, so combining components can occasionally add noise on small benchmarks. On REDDIT-B, the full vector (89.00) gains the most over any single block (Stats, 85.83).

\Cref{tab:sla_summary} summarizes the effect on rankings. The full vector has a higher average SLA (61.48) than any single block (56.88--59.99), but it is not strictly best: Spectral+Stats, at 28 dimensions and without PH, averages 61.69. Rankings barely depend on the choice. Every configuration keeps SLA rank correlation of at least 0.968 and AlignAvg rank correlation of at least 0.982 with the default. The AlignAvg figures are partly mechanical, since half of AlignAvg (centroid FLA) is held fixed. We keep the full vector as the default because it covers all three families, not because it dominates them.

\paragraph{Minimal descriptor.} \Cref{tab:minimal_descriptor} ranks the 48 dimensions by mutual information with the label, averaged across datasets. A 12-dimensional subset with no PH features (all 8 spectral dimensions and 4 statistics) reaches a rank correlation of 0.988 with the full vector. Larger subsets add little to the ranking and do not improve regime agreement. Since selection and evaluation use the same datasets, this is an optimistic estimate and a starting point for a lightweight variant rather than a validated replacement.

\begin{table}[t]
\centering
\caption{\footnotesize Minimal structural descriptor: top-$k$ dimensions by mutual information, averaged across datasets. Lap and Adj are the Laplacian and adjacency spectral blocks. Dimensions are selected and evaluated on the same 19 datasets, so these figures are in-sample.}
\label{tab:minimal_descriptor}
\footnotesize
\setlength{\tabcolsep}{5pt}
\begin{sc}
\begin{tabular}{clcccc}
\toprule
$k$ & \textbf{Blocks selected} & \textbf{Avg SLA} & \textbf{SLA $\rho$ vs.\ full} & \textbf{Pearson (GCN)} & \textbf{Regime agr.\ (\%)} \\
\midrule
8 & 3 Lap, 2 Adj, 3 Stats & 59.97 & 0.972 & 0.870 & 68.4 \\
12 & 4 Lap, 4 Adj, 4 Stats & 60.79 & 0.988 & 0.884 & 78.9 \\
16 & 4 PH, 4 Lap, 4 Adj, 4 Stats & 60.99 & 0.988 & 0.880 & 78.9 \\
24 & 8 PH, 4 Lap, 4 Adj, 8 Stats & 61.71 & 0.989 & 0.888 & 78.9 \\
\bottomrule
\end{tabular}
\end{sc}
\end{table}

\subsection{Probe Sufficiency Self-Check}\label{app:probe-sufficiency}
The reference alignment scores use a deliberately lightweight probe suite: logistic regression and RBF-kernel SVM. This design keeps SLA and FLA cheap, standardized, and independent of graph encoder training. A natural concern is probe underfitting: a low SLA or FLA score may occur because the restricted-view summary contains little accessible marginal label signal, or because the reference probes are not expressive enough to separate the summary vectors. We therefore introduce a simple probe sufficiency self-check.

\paragraph{Alternative probe suite.} We evaluate an alternative, more expressive probe suite consisting of five classifiers: logistic regression, linear SVM, RBF-SVM, a shallow MLP with two hidden layers and ReLU activations, and CatBoost. The same 5-seed, 10-fold stratified cross-validation protocol is used. This alternative suite is not intended to replace the reference protocol, but to test whether the dataset placement in the SLA-FLA plane is stable under a richer family of shallow predictors.

Let $\widehat{\mathrm{SLA}}_{\mathrm{ref}}$ and $\widehat{\mathrm{SLA}}_{\mathrm{alt}}$ denote the reference and alternative-probe SLA scores, respectively. We define the probe-sensitivity gap
\[\delta_{\mathrm{SLA}} = \widehat{\mathrm{SLA}}_{\mathrm{alt}} - \widehat{\mathrm{SLA}}_{\mathrm{ref}},\]
and analogously define $\delta_{\mathrm{FLA}} = \widehat{\mathrm{FLA}}_{\mathrm{alt}} - \widehat{\mathrm{FLA}}_{\mathrm{ref}}$. Positive values indicate that the alternative suite finds more accessible marginal signal than the reference suite, while negative values indicate that the reference suite is at least as strong under this protocol.

\paragraph{Self-check interpretation.} We interpret the gap as a probe-sensitivity diagnostic, not as a formal statistical test:
\begin{itemize}[leftmargin=1.5em, itemsep=2pt]
  \item $|\delta| \leq 5$ points: the reference score is empirically stable under the alternative probe suite, suggesting that probe underfitting is unlikely to be the dominant explanation for the observed alignment score.
  \item $|\delta| > 5$ points: the reference score should be treated as probe-sensitive. In this case, the alternative-probe score should be reported alongside the reference score, and conclusions about low alignment should be made cautiously.
\end{itemize}
The 5-point threshold is a practical heuristic based on the observed gap distribution in Table~\ref{tab:probe_gap}. It is intended to flag clear probe-sensitivity cases rather than define a universal cutoff.

\paragraph{Results.} Table~\ref{tab:probe_gap} shows that the reference alignment scores are stable for most datasets. For 15 of the 17 datasets, $|\delta_{\mathrm{SLA}}| \leq 5$ points. Gaps in both directions are roughly equally common, confirming that the reference protocol does not systematically underestimate marginal accessibility: negative gaps, where the reference suite matches or exceeds the alternative, are as frequent as positive ones.

The clearest exception is \textsc{Letter-High}, with $\delta_{\mathrm{SLA}} = +11.43$ and $\delta_{\mathrm{FLA}} = +21.54$. This dataset is atypical in our suite: it contains small graphs with average size 4.67, 2-dimensional node features, and 15 classes. The large positive gap indicates that the reference probes underestimate marginal accessibility on this dataset, while the alternative suite can exploit additional fine-grained feature geometry. We therefore treat \textsc{Letter-High} as probe-sensitive rather than genuinely low-alignment under all shallow probes.

\begin{wraptable}{r}{2.5in}
\centering
\caption{%
  \footnotesize
  Probe sufficiency self-check: difference between alternative-probe and reference alignment scores. We report $\delta_{\mathrm{SLA}} = \widehat{\mathrm{SLA}}_{\mathrm{alt}} - \widehat{\mathrm{SLA}}_{\mathrm{ref}}$ and analogously for FLA. Most datasets show small, unsystematic differences. \textsc{Letter-High} is the clearest probe-sensitive case.
}
\label{tab:probe_gap}
\footnotesize
\setlength{\tabcolsep}{8pt}
\renewcommand{\arraystretch}{1.15}
\begin{sc}
\resizebox{\linewidth}{!}{%
\begin{tabular}{lcc}
\toprule
\textbf{Dataset} & $\delta_{\mathrm{SLA}}$ & $\delta_{\mathrm{FLA}}$ \\
\midrule
AIDS            & $+0.06$  & $-1.24$  \\
MOLTOX21        & $-3.13$  & $-2.92$  \\
MUTAG           & $-1.71$  & $+0.78$  \\
DD              & $-0.13$  & $-3.97$  \\
BZR             & $-7.16$  & $-11.07$ \\
PROTEINS\_full  & $-2.18$  & $-0.25$  \\
MOLHIV          & $-4.33$  & $-5.53$  \\
BACE            & $-1.57$  & $-4.60$  \\
PROTEINS        & $-2.14$  & $-4.68$  \\
COX2            & $+4.86$  & $-8.26$  \\
NCI1            & $-1.18$  & $-2.32$  \\
FRANKENSTEIN    & $-0.14$  & $-2.59$  \\
ENZYMES         & $-2.29$  & $+5.57$  \\
MNIST           & $-0.10$  & $-0.43$  \\
CIFAR-10        & $+0.35$  & $-1.78$  \\
COIL-DEL        & $-2.10$  & $-4.29$  \\
\midrule
\textsc{Letter-High} & $\mathbf{+11.43}$ & $\mathbf{+21.54}$ \\
\bottomrule
\end{tabular}}
\end{sc}
\vspace{-0.2in}
\end{wraptable}

This interpretation is also consistent with the observed backbone spread on \textsc{Letter-High}: architectures with more expressive feature and attention mechanisms, such as SAGE and GPS, achieve substantially higher accuracy than GCN. Thus, a large probe gap is itself informative: it flags datasets where marginal signal may exist but requires more expressive access than the reference probe provides.

\paragraph{Practical guidance.} The self-check adds one lightweight step to the alignment pipeline. After computing the reference SLA and FLA scores, practitioners can run one or more stronger shallow probes, such as an MLP or CatBoost, on the same restricted-view summaries. If the resulting gap is small, the reference score can be treated as stable. If the gap is large, the dataset should be flagged as probe-sensitive, and both the reference and alternative-probe scores should be reported.

We emphasize that this self-check does not change the interpretation of SLA and FLA. They remain restricted-view diagnostics of marginal label accessibility under specified summary maps and probe families. The self-check simply guards against a known edge case: a low reference score caused by insufficient probe expressivity rather than weak signal in the restricted-view summary.

\paragraph{Individual probes.} \Cref{tab:probe_summary} scores each probe separately on the same summaries. Rank correlation with LogReg stays between 0.909 and 1.000. Pearson correlation with GCN accuracy stays between 0.877 and 0.926 for SLA. For FLA it stays between 0.911 and 0.931, except CatBoost at 0.789, which raises FLA sharply on ENZYMES and COLORS-3 relative to the other probes.

\begin{table}[t]
\centering
\caption{\footnotesize Individual probes on the same summaries. $\rho$ and regime agreement are measured against LogReg alone, not against the two-probe reference average. SLA uses 19 datasets and FLA uses 14.}
\label{tab:probe_summary}
\footnotesize
\setlength{\tabcolsep}{5pt}
\begin{sc}
\begin{tabular}{l|ccc|ccc}
\toprule
& \multicolumn{3}{c|}{\textbf{SLA}} & \multicolumn{3}{c}{\textbf{FLA}} \\
\textbf{Probe} & $\rho$ & \textbf{Pearson (GCN)} & \textbf{Regime (\%)} & $\rho$ & \textbf{Pearson (GCN)} & \textbf{Regime (\%)} \\
\midrule
LogReg & 1.000 & 0.880 & 100.0 & 1.000 & 0.931 & 100.0 \\
Linear SVM & 0.989 & 0.892 & 84.2 & 1.000 & 0.925 & 100.0 \\
RBF-SVM & 0.989 & 0.877 & 84.2 & 0.947 & 0.930 & 85.7 \\
MLP & 0.935 & 0.918 & 73.7 & 0.925 & 0.911 & 78.6 \\
CatBoost & 0.909 & 0.926 & 78.9 & 0.930 & 0.789 & 71.4 \\
\bottomrule
\end{tabular}
\end{sc}
\end{table}

\subsection{Feature Summary Robustness}\label{app:fla-ablation}
Centroid pooling discards higher-order information about the node-feature distribution, so an underpowered summary could create or enlarge the low-alignment regime. We therefore recompute FLA under four richer permutation-invariant summaries: per-feature mean and standard deviation; mean, standard deviation, and skewness; quantile pooling at 10/50/90\%; and the top-4 eigenvalues of the feature covariance. The probe protocol is unchanged.

\begin{table}[t]
\centering
\caption{\footnotesize FLA (\%) under five feature summaries on the 14 TU benchmarks with node attributes. \textbf{Bold} is the best score per row.}
\label{tab:fla_ablation}
\footnotesize
\setlength{\tabcolsep}{5pt}
\begin{sc}
\resizebox{\textwidth}{!}{%
\begin{tabular}{l|ccccc}
\toprule
\textbf{Dataset} & \textbf{Centroid} & \textbf{Mean+Std} & \textbf{Mean+Std+Skew} & \textbf{Quantile (10/50/90)} & \textbf{Cov.\ eigs (top-4)} \\
\midrule
AIDS & 96.86 & 96.12 & \textbf{97.28} & 94.74 & 77.69 \\
MUTAG & 69.58 & 67.58 & 66.42 & 67.44 & \textbf{69.69} \\
DD & 72.51 & \textbf{77.14} & 76.58 & 69.50 & 65.46 \\
BZR & 71.79 & 75.10 & \textbf{77.00} & 75.58 & 64.53 \\
PROTEINS\_full & 74.08 & \textbf{74.38} & 74.27 & 74.10 & 71.46 \\
PROTEINS & 64.02 & \textbf{71.81} & 71.67 & 70.36 & 67.88 \\
COX2 & 67.13 & 73.75 & \textbf{74.33} & 71.76 & 63.02 \\
NCI1 & 59.81 & 63.55 & \textbf{64.16} & 57.80 & 59.55 \\
FRANKENSTEIN & 54.66 & 55.15 & \textbf{56.66} & 55.92 & 55.90 \\
Synthie & \textbf{45.83} & 44.10 & 44.22 & 41.40 & 29.38 \\
ENZYMES & 36.38 & 45.12 & \textbf{51.10} & 43.60 & 24.68 \\
Letter-High & 23.27 & 33.50 & \textbf{37.36} & 34.24 & 23.46 \\
COLORS-3 & 24.22 & 28.65 & \textbf{28.82} & 22.30 & 26.50 \\
COIL-DEL & 5.70 & 12.28 & \textbf{21.48} & 18.01 & 5.06 \\
\midrule
\textbf{Average} & 54.70 & 58.45 & \textbf{60.10} & 56.91 & 50.30 \\
\bottomrule
\end{tabular}}
\end{sc}
\end{table}

\begin{table}[t]
\centering
\caption{\footnotesize Feature-summary robustness on 14 datasets. Rank correlations are Spearman against centroid FLA, for FLA and for AlignAvg (with full SLA held fixed). Regime agreement uses per-axis tertiles of centroid FLA.}
\label{tab:fla_summary}
\footnotesize
\setlength{\tabcolsep}{5pt}
\begin{sc}
\resizebox{\textwidth}{!}{%
\begin{tabular}{lcccccc}
\toprule
\textbf{Summary} & \textbf{Dim} & \textbf{Avg FLA} & \textbf{FLA $\rho$} & \textbf{AlignAvg $\rho$} & \textbf{Pearson (GCN)} & \textbf{Regime agr.\ (\%)} \\
\midrule
Centroid & $d$ & 54.70 & 1.000 & 1.000 & 0.934 & 100.0 \\
Mean+Std & $2d$ & 58.45 & 0.965 & 1.000 & 0.920 & 78.6 \\
Mean+Std+Skew & $3d$ & 60.10 & 0.943 & 0.978 & 0.925 & 71.4 \\
Quantile (10/50/90) & $3d$ & 56.91 & 0.934 & 0.982 & 0.936 & 78.6 \\
Cov.\ eigs (top-4) & $4$ & 50.30 & 0.947 & 0.978 & 0.842 & 85.7 \\
\bottomrule
\end{tabular}}
\end{sc}
\end{table}

\paragraph{Findings.} Richer summaries do raise FLA: average FLA rises from 54.70 for the centroid to 60.10 for mean+std+skew (\Cref{tab:fla_ablation}), so the centroid is not the highest-scoring summary. The effect is concentrated on some low-alignment datasets. Under mean+std+skew, COIL-DEL moves from 5.70 to 21.48, Letter-High from 23.27 to 37.36, and ENZYMES from 36.38 to 51.10. Low centroid FLA on these datasets is therefore partly a property of the summary, and we read it as a lower bound on feature accessibility.

The ordering and the relation to accuracy are nevertheless stable (\Cref{tab:fla_summary}). Rank correlation with centroid FLA is at least 0.934, and AlignAvg rank correlation is at least 0.978. Pearson correlation with GCN accuracy is 0.920--0.936 for mean+std, mean+std+skew, and quantile pooling, against 0.934 for the centroid; it falls to 0.842 for covariance eigenvalues, which discard the mean entirely. We keep the centroid as the default for its cost and simplicity. We do not evaluate trained summaries such as Deep Sets, because a summary fitted to labels would reintroduce the encoder training the diagnostic is designed to avoid.

\subsection{Regime and Correlation Stability}\label{app:regime-stability}
We now test whether the central conclusions, the low-alignment regime and the backbone-agnostic trend, survive these changes jointly. \Cref{tab:regime_stability} assigns each dataset a joint regime using fixed cutoffs and reports how many datasets change regime under each construction, together with the AlignAvg--accuracy correlation for all four backbones. The cutoffs are a reporting convention, since regime labels are relative to the chart that defines them.

\begin{table}[t]
\centering
\caption{\footnotesize Stability of joint regimes and of the backbone trend. A dataset is high/high if SLA and FLA are both $\geq 60$, low/low if both are $< 40$, and mixed otherwise. \textbf{Changed} counts datasets whose joint regime differs from the reference row of the same block. \textbf{HH$\leftrightarrow$LL} counts direct jumps between the two extremes. The last four columns are Pearson correlations between AlignAvg and accuracy. All rows above the last block use the 14 TU benchmarks with node attributes.}
\label{tab:regime_stability}
\footnotesize
\setlength{\tabcolsep}{5pt}
\begin{sc}
\resizebox{\textwidth}{!}{%
\begin{tabular}{l|cc|cccc}
\toprule
\textbf{Construction} & \textbf{Changed} & \textbf{HH$\leftrightarrow$LL} & \textbf{GCN} & \textbf{SAGE} & \textbf{GAT} & \textbf{GPS} \\
\midrule
Reference (full SLA, centroid FLA) & 0/14 & 0 & 0.939 & 0.709 & 0.970 & 0.775 \\
\midrule
FLA: Mean+Std & 2/14 & 0 & 0.935 & 0.715 & 0.973 & 0.799 \\
FLA: Mean+Std+Skew & 2/14 & 0 & 0.937 & 0.728 & 0.973 & 0.814 \\
FLA: Quantile (10/50/90) & 1/14 & 0 & 0.938 & 0.735 & 0.972 & 0.817 \\
FLA: Cov.\ eigs (top-4) & 0/14 & 0 & 0.898 & 0.663 & 0.955 & 0.767 \\
\midrule
SLA: PH & 1/14 & 0 & 0.928 & 0.692 & 0.952 & 0.740 \\
SLA: Spectral & 0/14 & 0 & 0.944 & 0.719 & 0.968 & 0.772 \\
SLA: Stats & 1/14 & 0 & 0.924 & 0.699 & 0.950 & 0.758 \\
SLA: PH+Spectral & 1/14 & 0 & 0.942 & 0.717 & 0.964 & 0.767 \\
SLA: PH+Stats & 1/14 & 0 & 0.926 & 0.700 & 0.956 & 0.763 \\
SLA: Spectral+Stats & 0/14 & 0 & 0.941 & 0.714 & 0.967 & 0.770 \\
\midrule
Probe: LogReg & 2/14 & 0 & 0.941 & 0.711 & 0.969 & 0.773 \\
Probe: Linear SVM & 2/14 & 0 & 0.945 & 0.706 & 0.973 & 0.769 \\
Probe: RBF-SVM & 2/14 & 0 & 0.935 & 0.707 & 0.969 & 0.775 \\
Probe: MLP & 2/14 & 0 & 0.949 & 0.693 & 0.989 & 0.768 \\
Probe: CatBoost & 3/14 & 0 & 0.919 & 0.611 & 0.978 & 0.713 \\
\midrule
Five-probe suite (\Cref{tab:probe_gap}), 17 datasets, reference & 0/17 & 0 & 0.955 & 0.764 & 0.957 & 0.789 \\
Five-probe suite (\Cref{tab:probe_gap}), 17 datasets, alternative & 5/17 & 0 & 0.925 & 0.812 & 0.916 & 0.849 \\
\bottomrule
\end{tabular}}
\end{sc}
\end{table}

\paragraph{Findings.} No dataset jumps directly between high/high and low/low under any construction. Every change is a single step into or out of the mixed regime, and at most 5 of 17 datasets change under any one construction. Labels near the cutoffs do move: ENZYMES and COX2 change most often. The per-axis tertile regimes in \Cref{tab:sla_summary,tab:fla_summary,tab:probe_summary} show the same behaviour with no direct High--Low jumps, and agreement of 71--90\% for non-reference constructions reflects moves to and from the middle tier.

\begin{figure*}[t]
  \centering
  \includegraphics[width=\textwidth]{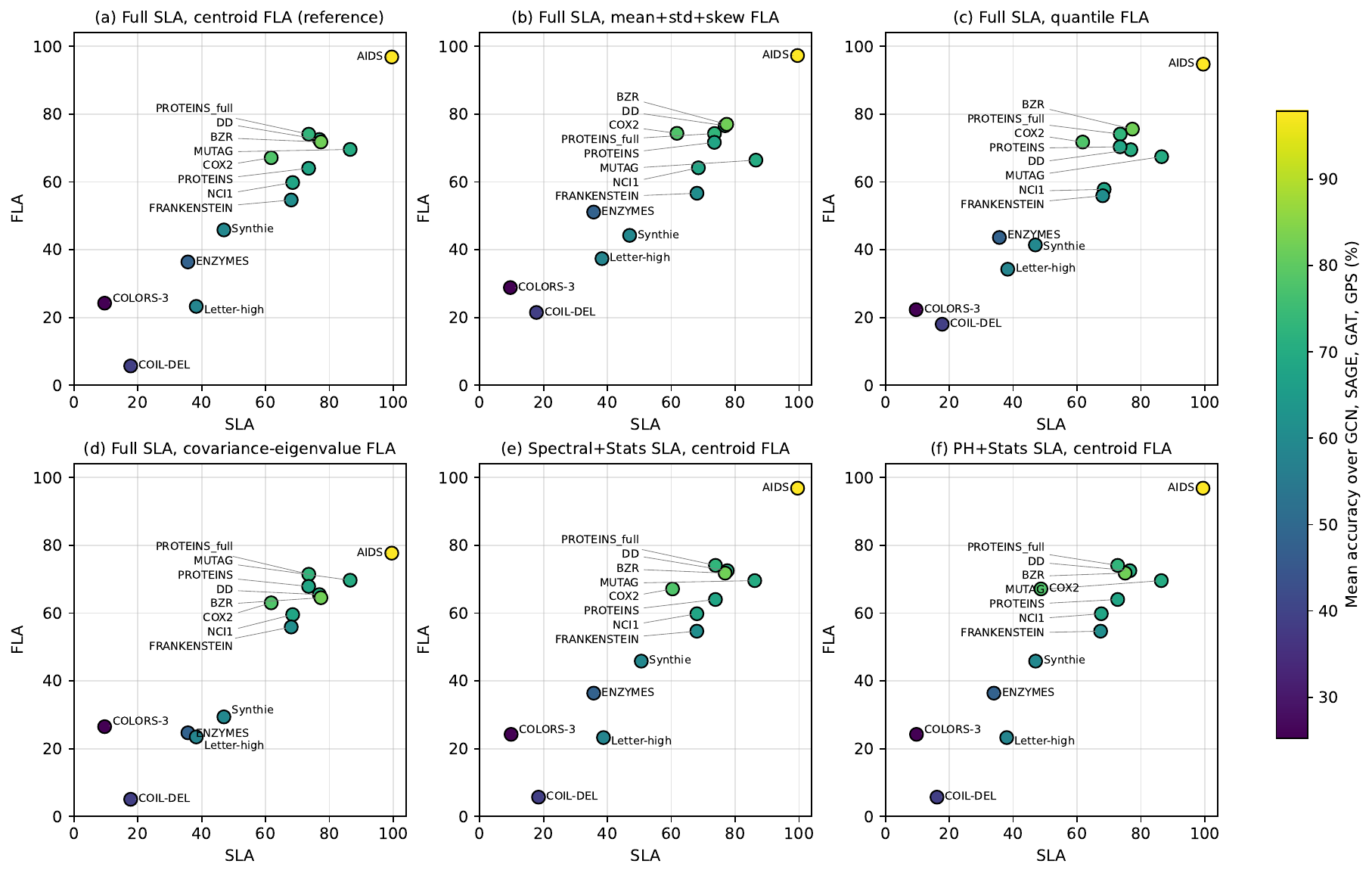}
  \caption{\textbf{The SLA--FLA plane under alternative constructions.} The 14 TU benchmarks with node attributes, colored by mean accuracy over the four backbones. Panel (a) is the reference construction; panels (b)--(d) change the feature summary and panels (e)--(f) change the structural blocks.}
  \label{fig:construction-robustness}
\end{figure*}

The backbone trend also holds. Across the 14-dataset rows, correlations stay within 0.898--0.989 for GCN and GAT and within 0.611--0.817 for SAGE and GPS, so the weaker relationship for SAGE and GPS is itself stable. In the five-probe plane on 17 datasets, correlations move from 0.955/0.764/0.957/0.789 to 0.925/0.812/0.916/0.849 (GCN/SAGE/GAT/GPS). The shifts run in both directions and neither plane is uniformly better. \Cref{fig:construction-robustness} shows that the layout of the plane is visually preserved under alternative constructions.

\subsection{Seed and Split Stability}\label{app:seed-stability}
We recompute SLA and FLA under 5 seeds, each drawing new 10-fold stratified splits and refitting the probes (\Cref{tab:seed_stability}). Correlations with GCN accuracy barely move, and the average per-dataset standard deviation across both views is 0.38 accuracy points, with a maximum of 1.15 (FLA on Synthie). This test varies the alignment scores only, since GCN accuracies are held fixed. It also covers random in-distribution splits only; \Cref{app:scaffold} examines scaffold splits, where the picture differs.

\begin{table}[t]
\centering
\caption{\footnotesize Stability across 5 seeds, each drawing new 10-fold splits and refitting the probes. GCN accuracies are held fixed at their \Cref{tab:gnn_baselines_extended} values. The last two columns are the mean and largest per-dataset standard deviation of the score, in accuracy points.}
\label{tab:seed_stability}
\footnotesize
\setlength{\tabcolsep}{5pt}
\begin{sc}
\begin{tabular}{lccccc}
\toprule
\textbf{View} & \textbf{Datasets} & \textbf{Pearson (GCN)} & \textbf{Spearman (GCN)} & \textbf{Mean std} & \textbf{Max std} \\
\midrule
SLA & 19 & 0.880 $\pm$ 0.002 & 0.842 $\pm$ 0.012 & 0.33 & 1.07 \\
FLA & 14 & 0.934 $\pm$ 0.003 & 0.889 $\pm$ 0.008 & 0.45 & 1.15 \\
\bottomrule
\end{tabular}
\end{sc}
\end{table}

\section{Heterophilic Ego-Networks}\label{app:ego_details}

\begin{table*}[ht]
    \centering
    \caption{Dataset statistics for Node-to-Graph heterophilic ego-graph benchmarks at varying $k$-hop sizes.}
    \label{tab:webkb_actor_statistics}
    \footnotesize
    \begin{sc}
    \resizebox{.8\textwidth}{!}{%
    \begin{tabular}{lcccccc}
    \toprule
    \textbf{Dataset} & $k$ & \textbf{\#Graphs} & \textbf{Av.Nodes} & \textbf{Av.Edges} & \textbf{\# Feat.} & \textbf{\#Classes} \\
    \midrule
    Texas & 1      & 183  & $4.05$    & $9.83$   & 1704 & 5 \\
    Texas & 2      & 183  & $66.68$  & $215.73$ & 1704 & 5 \\
    Texas & 3      & 183  & $127.83$ & $450.69$ & 1704 & 5 \\
    \midrule
    Cornell & 1   & 183  & $4.03$    & $8.74$   & 1704 & 5 \\
    Cornell & 2   & 183  & $55.86$  & $146.09$ & 1704 & 5 \\
    Cornell & 3   & 183  & $116.04$ & $360.03$ & 1704 & 5 \\
    \midrule
    Wisconsin & 1  & 251  & $4.59$    & $11.82$  & 1704 & 5 \\
    Wisconsin & 2  & 251  & $69.76$  & $245.19$ & 1704 & 5 \\
    Wisconsin & 3  & 251  & $158.31$ & $637.61$ & 1704 & 5 \\
    \midrule
    Actor & 1      & 7600 & $7.88$   & $22.23$  & 933  & 5 \\
    Actor & 2      & 7600 & $121.82$ & $374.78$ & 933  & 5 \\
    Actor & 3      & 7600 & $227.979$ & $379.08$ & 933  & 5 \\
    \bottomrule
    \end{tabular}
    }
    \end{sc}
\end{table*}

Heterophilic benchmarks test a different failure mode than the synthetic Twins: rather than suppressing marginal shortcuts by construction, they suppress them through label-structure anticorrelation. We convert four standard heterophilic node-classification graphs---Texas, Cornell, Wisconsin (WebKB), and Actor---into graph classification tasks by extracting $k$-hop ego-subgraphs and using the center-node label as the graph label, with a center-node indicator feature included in all experiments. We vary $k \in \{1,2,3\}$ to probe how much higher-order structure is needed before alignment and accuracy become recoverable. Dataset statistics are in \cref{tab:webkb_actor_statistics}.

\paragraph{Why heterophilic datasets require separate treatment.} In heterophilic regimes, immediate neighborhoods are label-adversarial by construction: 1-hop aggregation actively mixes signal from nodes of different classes. Increasing $k$ exposes longer-range structure that may restore label signal, but at the cost of incorporating more label-misaligned context. This bias--variance tradeoff is fundamentally different from the homophilic case and makes heterophilic ego-graphs a stress test for whether alignment scores track accuracy when the source of label signal is relational rather than marginal.

\subsection{Results}\label{app:ego_results}

\begin{table*}[t]
\centering
\caption{\footnotesize Pooled-model accuracy (\%) tracked against the alignment plane for heterophilic ego-graph datasets at varying hop sizes $k$. AlignAvg summarizes alignment as $(\mathrm{SLA}+\mathrm{FLA})/2$.}
\label{tab:webkb_actor_khop}
\footnotesize
\setlength{\tabcolsep}{5pt}
\begin{sc}
\resizebox{\textwidth}{!}{%
\begin{tabular}{l c|cccc|ccc}
\toprule
\textbf{Dataset} & \textbf{$k$} 
& \textbf{GCN} & \textbf{SAGE} & \textbf{GAT} & \textbf{GPS}
& \textbf{SLA} & \textbf{FLA} & \textbf{AlignAvg} \\
\midrule
Texas & 1
& 31.05\std{10.84} & 42.11\std{6.86} & 35.79\std{15.49} & 42.11\std{6.86}
& 33.78 & 43.16 & 38.47 \\
Texas & 2
& 35.26\std{4.88} & 36.32\std{6.32} & 25.79\std{13.87} & 36.32\std{6.09}
& 51.81 & 61.96 & 56.89 \\
Texas & 3
& 30.00\std{4.59} & 35.79\std{7.37} & 22.63\std{12.74} & 37.89\std{4.88}
& 62.67 & 61.84 & 62.26 \\
\midrule
Cornell & 1
& 36.32\std{3.87} & 35.26\std{12.85} & 35.26\std{6.36} & 60.00\std{7.88}
& 17.76 & 31.34 & 24.55 \\
Cornell & 2
& 27.89\std{4.88} & 25.79\std{1.97} & 22.11\std{3.57} & 58.95\std{2.68}
& 29.13 & 32.29 & 30.71 \\
Cornell & 3
& 27.37\std{6.78} & 33.68\std{9.47} & 20.53\std{3.07} & 58.95\std{1.29}
& 28.89 & 32.18 & 30.54 \\
\midrule
Wisconsin & 1
& 39.22\std{3.28} & 41.57\std{4.54} & 32.16\std{4.90} & 52.16\std{6.97}
& 27.86 & 44.72 & 36.29 \\
Wisconsin & 2
& 48.24\std{10.49} & 37.65\std{8.36} & 33.33\std{6.44} & 53.73\std{2.00}
& 37.99 & 45.45 & 41.72 \\
Wisconsin & 3
& 39.61\std{5.02} & 41.96\std{10.27} & 34.90\std{8.72} & 50.98\std{6.20}
& 34.93 & 46.08 & 40.51 \\
\midrule
Actor & 1
& 23.37\std{0.45} & 27.30\std{0.69} & 21.89\std{0.49} & 27.20\std{1.14}
& 18.20 & 24.12 & 21.16 \\
Actor & 2
& 22.76\std{0.73} & 25.08\std{0.18} & 19.99\std{0.74} & 25.75\std{0.36}
& 19.81 & 20.05 & 19.93 \\
Actor & 3
& 20.80\std{1.03} & 23.95\std{1.23} & 18.71\std{1.57} & 26.22\std{1.29}
& 9.99 & 19.87 & 14.93 \\
\bottomrule
\end{tabular}}
\end{sc}
\end{table*}

Results are summarized in \cref{tab:webkb_actor_khop}. The headline finding is that all pooled message-passing baselines achieve low accuracy across all datasets and hop sizes (typically 22--48\%, only modestly above 20\% chance for five classes), consistent with the low-alignment characterization. However, two patterns within this overall failure are worth isolating.

\paragraph{SLA rises with $k$ but accuracy does not follow.} Across all three WebKB datasets, structural alignment increases monotonically with hop radius---Texas rises from SLA $= 33.78$ at $k=1$ to $62.67$ at $k=3$, with Cornell and Wisconsin showing smaller but consistent gains. FLA remains stable throughout, confirming that the additional signal is structural rather than feature-driven. Yet accuracy peaks at $k=1$ or $k=2$ for most backbones and degrades at $k=3$. This dissociation between alignment and accuracy reveals a limitation of SLA as a predictor in heterophilic settings: our structural probe is permutation-invariant and captures global topology, but the discriminative signal at larger $k$ is increasingly entangled with label-misaligned context that pooled readouts cannot filter. Higher alignment does not help when the model cannot exploit it.

\paragraph{GPS is anomalously strong at small $k$.} On Cornell at $k=1$, GPS achieves $60.0\%$ accuracy despite very low alignment (SLA $= 17.76$, FLA $= 31.34$)---a 25-point gap over GCN and a clear outlier relative to the alignment-prediction story. This is not noise: the same pattern holds on Texas and Wisconsin at $k=1$. The likely explanation is that GPS's global attention and positional encodings allow it to exploit center-node identity and sparse structural cues that local MPNN aggregation cannot access when neighborhoods are adversarial. This constitutes a genuine scope boundary for the GLA predictor: in heterophilic settings at small $k$, architecture choice matters beyond what alignment scores capture, and GPS is the appropriate default.

\paragraph{Actor: alignment destruction under ego expansion.} Actor is the cleanest case in this section. SLA and FLA are uniformly low across all hop sizes (AlignAvg drops from $21.16$ at $k=1$ to $14.93$ at $k=3$), and accuracy decreases monotonically with $k$ for every backbone except GPS. Ego-graph expansion does not recover signal here---it destroys it, by diluting discriminative node features with structurally irrelevant context. Actor is a feature-dominated heterophilic regime where the center node's features carry the label signal and any aggregation beyond $k=1$ makes the task strictly harder.

\paragraph{Summary.} Heterophilic ego-graphs occupy a regime where the standard alignment--accuracy relationship breaks down in two distinct ways: SLA can rise while accuracy falls (WebKB, large $k$), and GPS can outperform its alignment score by a large margin (WebKB, small $k$). Both patterns point to the same underlying issue---pooled readouts cannot selectively aggregate label-aligned structure when label-misaligned context dominates the ego-graph. The practical implication is concrete: in heterophilic node-to-graph transfer, alignment scores remain useful for ruling out hopeless configurations (Actor at $k=3$) but are insufficient as predictors of achievable accuracy; backbone choice and ego-graph radius must be tuned jointly.

\section{Synthetic Sweep Experiments} \label{app:synthetic}
This appendix provides full implementation details for the synthetic alignment experiments described in \Cref{sec:synthetic}. Its purpose is to ensure reproducibility, clarify how alignment scores behave under controlled interventions, and formally justify the diagnostic role of the Twins construction.

\begin{figure*}[t]
  \begin{center}
    \centerline{
        \includegraphics[width=0.5\columnwidth]{figures/synthetic/synth_gcn.pdf} \includegraphics[width=0.5\columnwidth]{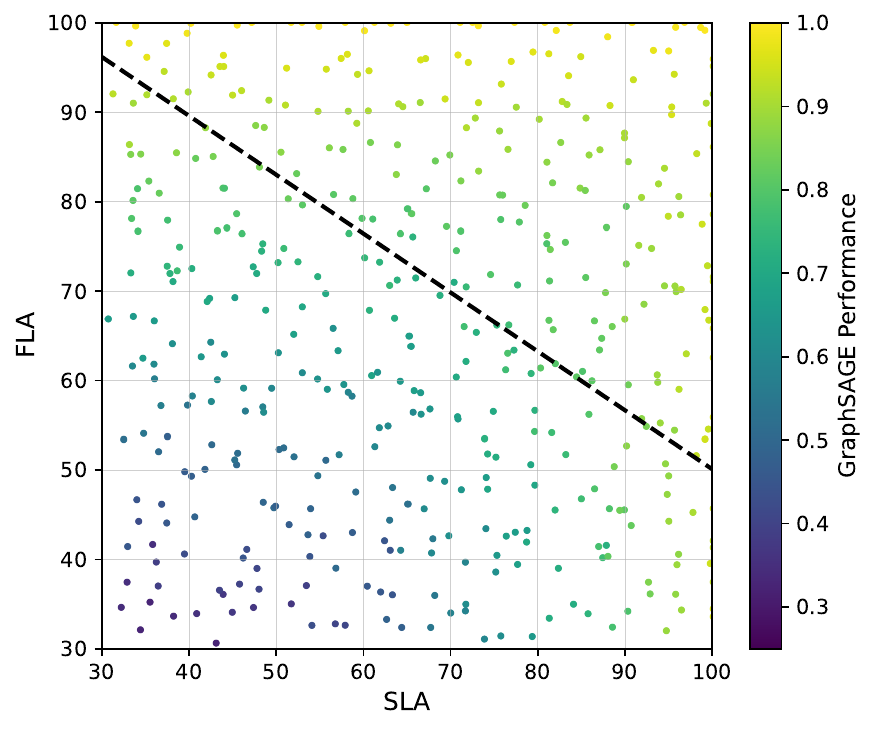}
    }\centerline{
        \includegraphics[width=0.5\columnwidth]{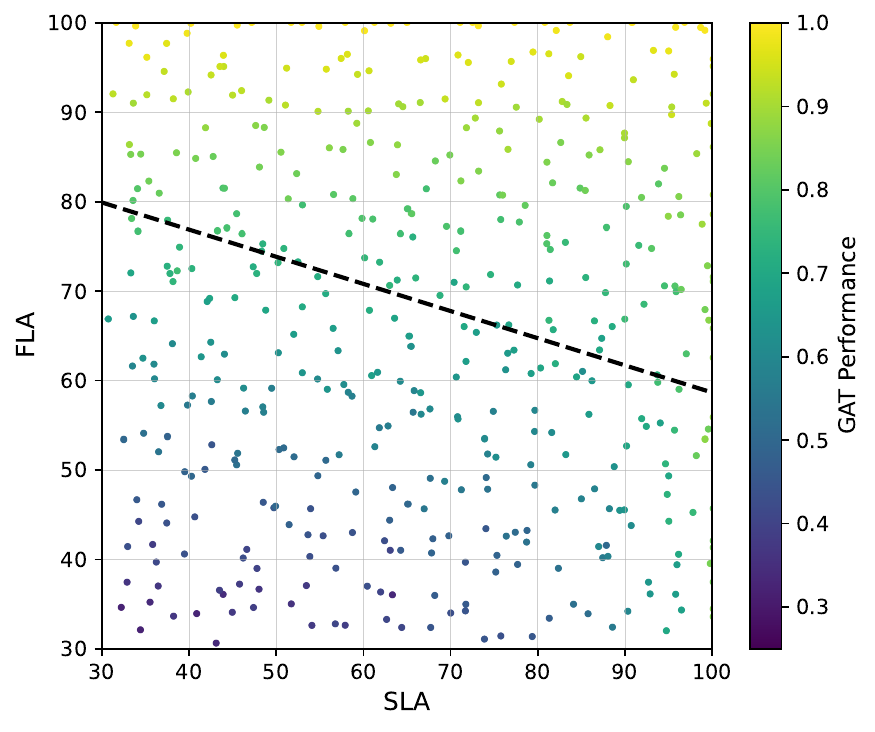} \includegraphics[width=0.5\columnwidth]{figures/synthetic/synth_gps.pdf}
    }
    \caption{\textbf{GNN performance across synthetic sweep with boundaries.} For a full synthetic sweep across the SLA-FLA plane colored by GNN accuracy. The dashed line marks a decision boundary for ``good'' performance set at 80\% accuracy. We include results for all four backbones, beginning in the top left: GCN, GraphSAGE, GAT, and GPS.}
    \label{fig:synth-plane}
  \end{center}
\end{figure*}

\subsection{Synthetic Dataset Generation Details}\label{app:synth_details}
All synthetic datasets are generated with fixed global statistics unless explicitly varied by sweep parameters. Each dataset consists of $N=1000$ graphs per sweep point. Graph sizes are sampled independently as $n \sim \mathcal{N}(120, 20^2)$ and clipped to $[60,200]$. Node feature dimension is fixed to $d=16$.

For alignment plane sweeps, the structural separation parameter $s \in [0,1]$ and feature separation parameter $f \in [0,1]$ are discretized on a uniform $15 \times 15$ grid. Each grid point corresponds to an independently generated dataset, with identical random seeds reused across all GNN backbones.

Unless otherwise stated, Gaussian noise variance is fixed to $\sigma^2=0.1$. Class mean vectors $\{\mu_y\}$ are sampled once per dataset family, normalized to unit norm, and scaled to a fixed signal magnitude.

\subsection{Structural Signal Control and SLA Behavior}\label{app:synth_sla}
Structural label alignment (SLA) is controlled exclusively by the structural separation parameter $s$. When $s=0$, all graphs are sampled from the shared generator $\mathcal{G}_0$, yielding identical structural distributions across classes and SLA near chance. As $s \to 1$, graphs are increasingly sampled from class-specific generators $\mathcal{G}_y$, which differ in global structural statistics such as degree distributions, clustering coefficients, and mesoscopic community structure.

Graph sizes and expected degrees are matched across classes, ensuring that increases in SLA are not driven by trivial size or density cues. Empirically, measured $\widehat{\mathrm{SLA}}$ increases monotonically with $s$ across all sweep families, validating $s$ as a calibrated control knob for structural label signal.

\subsection{Feature Signal Control and FLA Behavior}\label{app:synth_fla}
Feature label alignment (FLA) is controlled by the feature separation parameter $f$. When $f=0$, all node features are i.i.d.\ Gaussian noise, matched across labels, and pooled feature summaries are non-informative. As $f$ increases, a class-dependent mean shift $\mu_y$ is added to all nodes in graphs of label $y$, causing graph-level feature aggregates to become increasingly predictive.

Noise variance and feature dimensionality are fixed across all sweep points, so that $\widehat{\mathrm{FLA}}$ varies smoothly and monotonically with $f$. This construction isolates marginal feature signal: no structural information is required to achieve high FLA.

\subsection{Alignment Plane Coverage}\label{app:synth_plane}
Jointly sweeping $(s,f)$ induces datasets spanning the full alignment plane, including high-SLA/low-FLA, low-SLA/high-FLA, and jointly aligned regimes. Measured alignment scores $(\widehat{\mathrm{SLA}}, \widehat{\mathrm{FLA}})$ closely track the intended control parameters $(s,f)$, with small deviations attributable to finite-sample effects and injected noise.

We verify that sweep families occupy well-separated regions of the alignment plane and that no unintended correlations between SLA and FLA arise outside the jointly aligned regime. This confirms that the plane sweep provides a controlled testbed for analyzing how model performance organizes with respect to alignment.

All synthetic sweep output can be viewed in \cref{tab:synth_dataset} and \cref{fig:synth-plane}.

\begin{algorithm}[t]
\footnotesize
\caption{Twin synthetic graph generation (one graph)}
\label{alg:twin_gen}
\footnotesize
\begin{algorithmic}[1]
\STATE Sample SBM graph $G=(V,E)$ with fixed $(K,P)$ and $|V|=n$.
\STATE Compute node scores $s(v)$ using a chosen filtration (Linear, Radial, PageRank, Eigen).
\STATE Threshold $s(v)$ at empirical quantiles to form three groups $V_1,V_2,V_3$ (roughly balanced).
\STATE Assign features: for $v\in V_j$, set $x_v=\mu_j+\epsilon_v$, $\epsilon_v\sim\mathcal{N}(0,\sigma^2 I)$.
\STATE Output $(G,X)$ with class label determined by the assignment rule(s) (random vs.\ structured).
\end{algorithmic}
\end{algorithm}

\subsection{Twins Construction and Shortcut Elimination}\label{app:twins_details}
The Twins construction is designed to eliminate both structure-only and feature-only shortcuts. All classes share: (i) identical SBM parameters and identical edge sets per base graph, and (ii) identical multisets of node features.

As a result, classifiers operating solely on graph structure or on pooled feature statistics achieve near-chance performance. Labels differ only in the coupling between node roles and node feature types, yielding a strictly interaction-only regime.

\subsection{Role Scores and Balanced Assignments}\label{app:twins_roles}
For each base graph, nodes are assigned a scalar score $s(v)$ using one of four filtrations: \textbf{Linear}, \textbf{Radial}~\cite{seidman1983network}, \textbf{PageRank}~\cite{page1999pagerank}, or \textbf{Eigen}. The scores are then partitioned into three equal-sized groups according to the first and second tertiles of the score distribution. Feature prototypes $(\mu_1,\mu_2,\mu_3)$ are assigned deterministically to these groups, with additive Gaussian noise. This guarantees that all classes have identical prototype counts, ensuring that marginal feature distributions are exactly matched.

The algorithm for each of the scalar scoring functions is described below.

\paragraph{Linear.} Nodes are scored by shortest-path distance from a peripheral root. A degree-1 root is chosen when available, else the next smallest-degree node is used. This permutation yields distance-based stratification, with distinct feature prototypes occupying successive shortest-path shells.

\paragraph{Radial.} Nodes are scored by core number, defined as the largest $k$ such that the node belongs to a $k$-core, a maximal subgraph in which every node is of degree $k$ or greater. This induces core-based feature stratification, with the different feature prototypes occupying successive shells radiating outward from the densest regions of the graph.

\paragraph{PageRank.} Nodes are scored by PageRank, which computes the stationary probability of a random walker occupying a given node under a teleportation model. This produces global-importance stratification, with higher-scoring nodes tending to lie at well-connected positions that link multiple regions of the graph.

\paragraph{Eigen.} Nodes are scored by eigenvector centrality, which is calculated for each connected component and linearly normalized to $[0,1]$, with singletons assigned a zero score. This induces spectral-importance stratification in which high-score nodes connect to other high-score nodes, forming a per-component hub-periphery structure. This permutation scheme is similar to the radial distribution when dense cores are well connected, but eigenvector centrality captures a more global influence, whereas k-core only measures local subgraph density.

\subsection{Interpretation of Alignment Scores in Twins}\label{app:twins_alignment}
Although Twins datasets exhibit low measured SLA and FLA, this does not contradict lower-bound intuitions derived from marginal alignment. Both alignment scores are intentionally suppressed by construction; however, labels remain predictable through higher-order interactions between structure and features.

Alignment scores measure marginal separability under pooled summaries. Twins instead isolate a regime in which prediction depends on preserving global correspondence between node roles and feature types. This highlights a fundamental limitation of marginal alignment metrics and motivates evaluation beyond SLA and FLA alone.

\subsection{Binary and Multiclass Twins Settings}\label{app:twins_classes}
We evaluate both binary Twins (random vs.\ one structured assignment) and multiclass Twins (random vs.\ two structured rules, e.g., Linear vs.\ Radial). In multiclass settings, chance accuracy decreases accordingly, and successful prediction requires distinguishing multiple interaction patterns despite identical marginal distributions.

\section{Background on Topological and Spectral Signatures} \label{app:background}
\subsection{Persistent Homology}\label{sec:PH}
Persistent Homology (PH) is a central tool in Topological Data Analysis (TDA) that captures multiscale topological features such as connected components, loops, and voids that persist across different scales~\cite{dey2022computational}. Originally developed for point cloud data, PH has since been successfully extended to graphs, images, and other data modalities~\cite{coskunuzer2024topological}. In graph learning, PH provides latent structural insights that complement conventional methods and GNNs, enabling a richer understanding of both local and global graph structure.

\begin{wrapfigure}{r}{2.5in}
\centering
\includegraphics[width=\linewidth]{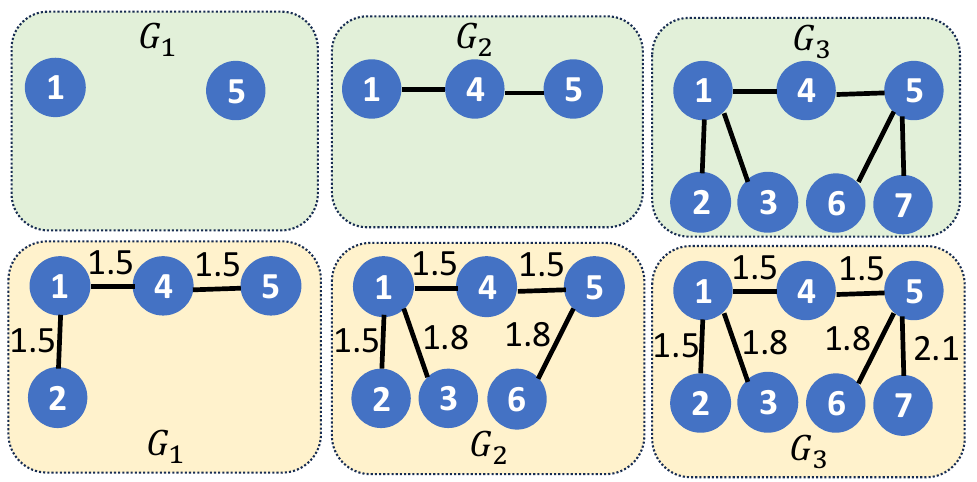}
\caption{\footnotesize Graph Filtration: For $\mathcal{G}=\mathcal{G}_3$ in both examples, the top figure illustrates a \textit{superlevel filtration using the node degree function} with thresholds $3>2>1$, where nodes of degree 3 are activated first, followed by those of lower degrees.  Similarly, the bottom figure illustrates a \textit{sublevel filtration based on edge weights} with thresholds $1.5< 1.8< 2.1$.  \label{fig:graph-filtration}}
\vspace{-.2in}
\end{wrapfigure}

PH typically consists of three main steps: constructing a filtration, obtaining persistence diagrams, and performing vectorization. The process begins with \textit{filtration}, where a nested sequence of simplicial complexes is built to track the evolution of topological features. Given an unweighted graph $\mathcal{G} = (\mathcal{V}, \mathcal{E})$ and a filtration function $f: \mathcal{V} \to \mathbb{R}$ with thresholds $\mathcal{I} = \{\alpha_i\}$, we construct subgraphs $\mathcal{G}^i = (\mathcal{V}_i, \mathcal{E}_i)$, where $\mathcal{V}_i = \{v \in \mathcal{V} \mid f(v) \leq \alpha_i\}$ (See~\Cref{fig:graph-filtration} for toy example). The filtration function can be based on graph structural properties (e.g., degree, betweenness centrality) or derived from domain-specific information (e.g., atomic number in molecular graphs). For weighted graphs, edge weights can also serve as a natural filtration~\cite{coskunuzer2024topological}. The primary goal is to establish a hierarchy among nodes or edges that yields a meaningful sequence of subgraphs aligned with the downstream task. In temporal graphs, a natural choice would be timestamps on edges as a valuable filtration function.

Once the subgraphs are defined, each $\mathcal{G}^i$ is extended to its clique complex $\widehat{\mathcal{G}}^i$, producing a sequence $\widehat{\mathcal{G}}^1 \subseteq \widehat{\mathcal{G}}^2 \subseteq \dots \subseteq \widehat{\mathcal{G}}^N$. As the filtration evolves, topological features such as connected components (0-holes), loops (1-holes), and cavities (2-holes) emerge and disappear. Each $k$-dimensional feature $\sigma$ is associated with a birth index $b_\sigma$ and death index $d_\sigma$ ($1 \leq b_\sigma < d_\sigma \leq N$), recorded as a point $(b_\sigma, d_\sigma)$ in the $k$-th persistence diagram ${\rm PD}_k(\mathcal{G}) = \{(b_\sigma, d_\sigma) \mid \sigma \in H_k(\widehat{\mathcal{G}}^i)\}$. Since persistence diagrams are sets of points in $\mathbb{R}^2$, a \textit{vectorization} step is necessary for compatibility with machine learning models. Techniques such as persistence images, landscapes, silhouettes, and Betti curves~\cite{ali2023survey} are commonly used to transform diagrams into fixed-size feature representations.

While PH effectively captures complex structural patterns, its integration into temporal graph learning remains challenging. Most prior methods are limited to static graphs and fail to model evolving topological features. Additionally, combining PH summaries with temporal dynamics and node attributes is nontrivial. To address these issues, we propose a hybrid framework that extracts topological descriptors from temporal subgraphs, aligning them with sliding windows to track structural changes over time. This enables scalable, interpretable temporal graph classification without relying on rigid snapshots or heavy discretization.

\subsection{Spectral Signatures for Graphs}\label{sec:spectral}
Spectral graph theory provides permutation-invariant summaries of graph structure through eigenvalues of matrices such as the adjacency matrix $A$ and the normalized Laplacian $L=I-D^{-1/2}AD^{-1/2}$~\cite{chung1997spectral,spielman2011spectral}. Since eigenvalues are invariant under node relabeling, they offer compact, canonical descriptors that reflect global connectivity patterns, expansion, and community structure.

In this work, we use a \emph{truncated spectrum} rather than a full density-of-states histogram. Concretely, our spectral vector concatenates (i) the four smallest nonzero eigenvalues of $L$ and (ii) the four largest eigenvalues of $A$ (see Appendix~\Cref{app:struct_summary}). The low end of the Laplacian spectrum captures coarse connectivity and bottlenecks: the multiplicity of the $0$ eigenvalue equals the number of connected components, and the first nonzero eigenvalue (algebraic connectivity) quantifies how well-knit the graph is, with larger values indicating fewer global bottlenecks~\cite{chung1997spectral}. The top adjacency eigenvalues capture dominant interaction scales and degree heterogeneity; for many graph families they correlate with the presence of hubs and overall edge concentration, providing a complementary view to Laplacian features~\cite{chung1997spectral}.

Using only a small number of eigenvalues yields a fixed-budget, low-dimensional summary that is easy to compute and compare across graphs of varying size, while still encoding informative global structure. This design matches our goal in constructing lightweight structural probes: the truncated spectral signature adds global structural sensitivity beyond local statistics without introducing high-dimensional histograms or model-specific embeddings.

\section{Per-Backbone Graph Label Alignment}\label{app:gla-recipe}
SLA and FLA scores are backbone-independent, using our synthetic sweep across the alignment atlas we can then target backbone-specific recipes to predict GNN performance by fitting a linear predictor
\[\textrm{GLA} = \alpha\cdot\textrm{SLA} + (1-\alpha)\cdot\textrm{FLA}\]
The optimal convex combinations across all backbones are
\begin{align*}
    &\textrm{GLA}_{\textrm{GCN}} = 0.45\textrm{SLA} + 0.55\textrm{FLA}, \quad &\textrm{GLA}_{\textrm{SAGE}} = 0.40\textrm{SLA} + 0.60\textrm{FLA}, \\
    &\textrm{GLA}_{\textrm{GAT}} = 0.23\textrm{SLA} + 0.77\textrm{FLA}, \quad
    &\textrm{GLA}_{\textrm{GPS\,\,\,}} = 0.09\textrm{SLA} + 0.91\textrm{FLA}.
\end{align*}

\subsection{Evaluation On Real Datasets}
Predicted performance, along with computed absolute differences, can be viewed in \cref{tab:gla_abs_diff}.

\begin{table*}[t]
\centering
\caption{\footnotesize Absolute deviation between predicted GLA scores and observed pooled accuracy. For each backbone, $\Delta_{\mathrm{BB}} = |\widehat{\mathrm{GLA}}_{\mathrm{BB}} - \mathrm{Acc}_{\mathrm{BB}}|$. Accuracies use reported means. Featureless datasets (FLA undefined) are omitted.}
\label{tab:gla_abs_diff}
\footnotesize
\setlength{\tabcolsep}{4.2pt}
\begin{sc}
\resizebox{\textwidth}{!}{%
\begin{tabular}{l|ccc|ccc|ccc|ccc}
\toprule
\textbf{Dataset}
& $\widehat{\mathrm{GLA}}_{\mathrm{GCN}}$ & \textbf{GCN} & $\Delta_{\mathrm{GCN}}$
& $\widehat{\mathrm{GLA}}_{\mathrm{SAGE}}$ & \textbf{SAGE} & $\Delta_{\mathrm{SAGE}}$
& $\widehat{\mathrm{GLA}}_{\mathrm{GAT}}$ & \textbf{GAT} & $\Delta_{\mathrm{GAT}}$
& $\widehat{\mathrm{GLA}}_{\mathrm{GPS}}$ & \textbf{GPS} & $\Delta_{\mathrm{GPS}}$ \\
\midrule
AIDS           & 98.05 & 98.40 & 0.35 & 97.91 & 98.30 & 0.39 & 97.46 & 95.95 & 1.51 & 97.09 & 98.85 & 1.76 \\
MOLTOX21       & 86.75 & 73.45 & 13.30 & 86.97 & 73.63 & 13.34 & 87.71 & 71.95 & 15.76 & 88.32 & 73.70 & 14.62 \\
MUTAG          & 77.32 & 71.28 & 6.04 & 76.47 & 69.74 & 6.73 & 73.57 & 67.18 & 6.39 & 71.18 & 70.77 & 0.41 \\
DD             & 74.41 & 70.30 & 4.11 & 74.19 & 70.55 & 3.64 & 73.45 & 69.03 & 4.42 & 72.84 & 71.90 & 0.94 \\
BZR            & 74.17 & 81.48 & 7.31 & 73.89 & 80.00 & 6.11 & 72.93 & 83.46 & 10.53 & 72.14 & 83.21 & 11.07 \\
PROTEINS\_full & 73.80 & 71.07 & 2.73 & 73.82 & 71.70 & 2.12 & 73.89 & 73.84 & 0.05 & 73.95 & 74.11 & 0.16 \\
MOLHIV         & 69.61 & 71.41 & 1.80 & 68.97 & 74.47 & 5.50 & 66.79 & 73.39 & 6.60 & 64.98 & 71.26 & 6.28 \\
BACE           & 68.60 & 72.76 & 4.16 & 68.35 & 77.17 & 8.82 & 67.49 & 60.53 & 6.96 & 66.78 & 69.74 & 2.96 \\
PROTEINS       & 68.28 & 67.59 & 0.69 & 67.80 & 68.12 & 0.32 & 66.17 & 67.05 & 0.88 & 64.85 & 68.93 & 4.08 \\
COX2           & 64.61 & 78.09 & 13.48 & 64.86 & 78.30 & 13.44 & 65.73 & 76.81 & 11.08 & 66.45 & 77.87 & 11.42 \\
NCI1           & 63.66 & 68.30 & 4.64 & 63.23 & 69.39 & 6.16 & 61.78 & 64.53 & 2.75 & 60.53 & 69.39 & 8.86 \\
FRANKENSTEIN   & 60.66 & 59.12 & 1.54 & 59.99 & 62.07 & 2.08 & 57.73 & 61.57 & 3.84 & 55.86 & 60.53 & 4.67 \\
Synthie        & 46.57 & 69.00 & 22.43 & 46.51 & 77.00 & 30.49 & 46.28 & 45.50 & 0.78 & 46.10 & 45.00 & 1.10 \\
ENZYMES        & 36.09 & 47.50 & 11.41 & 36.14 & 50.33 & 14.19 & 36.31 & 43.00 & 6.69 & 36.44 & 50.67 & 14.23 \\
Letter-High    & 29.96 & 41.38 & 11.42 & 29.21 & 79.29 & 50.08 & 26.65 & 30.80 & 4.15 & 24.36 & 81.51 & 57.15 \\
MNIST          & 24.77 & 37.85 & 13.08 & 24.67 & 48.80 & 24.13 & 24.34 & 32.01 & 7.67 & 24.07 & 47.33 & 23.26 \\
CIFAR-10       & 21.49 & 27.77 & 6.28 & 21.77 & 39.85 & 18.08 & 22.71 & 31.70 & 8.99 & 23.48 & 41.54 & 18.06 \\
COLORS-3       & 17.65 & 24.92 & 7.27 & 18.38 & 26.08 & 7.70 & 20.84 & 25.08 & 4.24 & 22.87 & 25.03 & 2.16 \\
COIL-DEL       & 11.10 & 33.97 & 22.87 & 10.50 & 55.59 & 45.09 & 8.44  & 18.85 & 10.41 & 6.75  & 46.08 & 39.33 \\
\bottomrule
\end{tabular}}
\end{sc}
\end{table*}

Across \textbf{molecular benchmarks}~\cite{morris2020tudataset}, deviations are consistently small, particularly for GAT and GPS, indicating that classification accuracy in this regime is well explained by alignment alone. These datasets are characterized by moderate graph sizes, low class counts (predominantly binary), and semantically meaningful node features, placing them squarely within the regime where SLA and FLA act as sufficient summary statistics of task difficulty. In this setting, GLA functions as a faithful, backbone-aware predictor of performance.

In contrast, \textbf{vision-style graph datasets} (e.g., MNIST, CIFAR-10, Letter-High, COIL-DEL) exhibit large and systematic deviations, with GLA substantially under-predicting observed accuracy, especially for feature-centric backbones. These datasets combine weak or artificially induced graph topology with low-dimensional but highly informative node features and higher class counts. Consequently, performance is dominated by raw feature geometry rather than graph-level alignment, violating the assumptions underlying GLA and revealing a clear failure mode of alignment-based predictors outside alignment-limited regimes.

The \textbf{synthetic benchmarks} provide controlled validation of backbone-specific inductive biases. In these settings, attention-based and global models (GAT and GPS) closely track GLA predictions, while classical message-passing architectures (GCN and GraphSAGE) exhibit larger deviations. This behavior aligns with the designed disentanglement of structural and feature cues in the synthetic data and confirms that the learned GLA weights meaningfully capture differences in backbone reliance on feature versus structural alignment.

Finally, for \textbf{social graph benchmarks}, node features are absent and FLA is undefined, rendering GLA inapplicable by construction. Performance in this regime is governed by higher-order structural statistics, graph size effects, and dataset-specific biases that are not captured by SLA alone. This delineates a clear boundary of applicability for alignment-based predictors and motivates the need for richer structural descriptors in featureless settings.

Overall, this analysis establishes GLA as a \emph{domain-conditional predictor} of GNN performance: effective in alignment-limited regimes with informative node features, but insufficient in domains where task signal arises from non-alignment cues or higher-order structure.

\begin{table}[t]
\centering
\caption{\footnotesize Practical backbone choice as a function of SLA/FLA regime.}
\label{tab:backbone_choice_rules}
\setlength{\tabcolsep}{6pt}
\renewcommand{\arraystretch}{1.15}
\resizebox{\textwidth}{!}{%
\begin{tabular}{ll|l|l}
\toprule
\textbf{SLA} & \textbf{FLA} & \textbf{Recommended backbone} & \textbf{Rationale} \\
\midrule
High & High & GCN/SAGE (cheapest) & Alignment is easy; capacity adds little, prefer efficiency. \\
High & Low/Med & GCN (or SAGE) & Structure-dominant; simpler MPNNs suffice and generalize well. \\
Med & Med & SAGE/GAT & Balanced regime; moderate capacity often best. \\
Low/Med & High & GPS (or GAT) & Feature-dominant; attention/global aggregation helps. \\
Low & Low & GPS/GAT (expect low) & Neither cue aligns; need capacity but accuracy remains limited. \\
\bottomrule
\end{tabular}}
\end{table}

\begin{sc}
\small
{\footnotesize\setlength{\tabcolsep}{5pt}
\begin{longtable}{l|cccc|ccc}
\caption{Synthetic sweep results reported in dataset-style format. All GNN results are reported as mean $\pm$ standard deviation (in percentage points), computed over multiple random seeds. Synthetic datasets were generated consistently across backbones using 1000 graphs per configuration, approximately 120 nodes ($\pm 20$ variance), feature dimension 16, and a $15\times15$ SLA/FLA grid. AlignAvg denotes the arithmetic mean of SLA and FLA (lower is worse).} \label{tab:synth_dataset} \\
\toprule
\textbf{Dataset} & \textbf{GCN} & \textbf{SAGE} & \textbf{GAT} & \textbf{GPS} & \textbf{SLA} & \textbf{FLA} & \textbf{AlignAvg} \\
\midrule
\endfirsthead
\caption[]{Synthetic sweep results reported in dataset-style format. All GNN results are reported as mean $\pm$ standard deviation (in percentage points), computed over multiple random seeds. Synthetic datasets were generated consistently across backbones using 1000 graphs per configuration, approximately 120 nodes ($\pm 20$ variance), feature dimension 16, and a $15\times15$ SLA/FLA grid. AlignAvg denotes the arithmetic mean of SLA and FLA (lower is worse).} \\
\toprule
\textbf{Dataset} & \textbf{GCN} & \textbf{SAGE} & \textbf{GAT} & \textbf{GPS} & \textbf{SLA} & \textbf{FLA} & \textbf{AlignAvg} \\
\midrule
\endhead
\midrule
\multicolumn{8}{r}{Continued on next page} \\
\midrule
\endfoot
\bottomrule
\endlastfoot
Synth-001 & 33.83\std{3.21} & 34.67\std{0.29} & 32.17\std{4.07} & 34.67\std{3.06} & 32.17 & 34.68 & 33.42 \\
Synth-002 & 30.33\std{3.21} & 32.83\std{3.75} & 32.17\std{5.39} & 29.50\std{2.29} & 34.37 & 32.17 & 33.27 \\
Synth-003 & 42.00\std{2.29} & 39.00\std{3.28} & 41.67\std{2.75} & 37.00\std{3.04} & 36.18 & 39.75 & 37.97 \\
Synth-004 & 37.33\std{4.48} & 35.00\std{2.00} & 36.67\std{2.52} & 38.83\std{4.54} & 32.83 & 37.50 & 35.17 \\
Synth-005 & 45.00\std{0.87} & 45.33\std{3.33} & 43.17\std{2.57} & 43.00\std{6.14} & 32.90 & 41.48 & 37.19 \\
Synth-006 & 42.00\std{5.89} & 42.50\std{3.00} & 41.00\std{4.27} & 43.83\std{1.89} & 34.17 & 44.30 & 39.23 \\
Synth-007 & 47.17\std{4.65} & 44.83\std{3.75} & 45.83\std{3.06} & 43.83\std{4.01} & 33.95 & 46.72 & 40.33 \\
Synth-008 & 45.33\std{1.61} & 44.67\std{1.61} & 45.50\std{1.00} & 43.50\std{4.58} & 36.80 & 46.22 & 41.51 \\
\end{longtable}}
\end{sc}

\section{Screening Use and Split Protocol}\label{app:screening}
This appendix evaluates alignment as a pre-training screen: what it costs, how well it predicts accuracy on unseen datasets, and what it says about benchmark quality. It then tests how far the results extend beyond random splits.

\subsection{Compute Cost}\label{app:runtime}
\Cref{tab:runtime} compares one pass of the alignment pipeline with one training run of each backbone, across a 22-fold range of dataset sizes on a single CPU-only machine.

\begin{table}[t]
\centering
\caption{\footnotesize Wall-clock seconds for one pass of GLA (one seed, 10-fold CV, LogReg and RBF-SVM on SLA and FLA) against one 100-epoch training run of each backbone with a single hyperparameter configuration, on one machine.}
\label{tab:runtime}
\footnotesize
\setlength{\tabcolsep}{5pt}
\begin{sc}
\resizebox{\textwidth}{!}{%
\begin{tabular}{lr|ccccc|cccc|cc}
\toprule
\textbf{Dataset} & \textbf{Graphs} & \textbf{SLA build} & \textbf{SLA CV} & \textbf{FLA build} & \textbf{FLA CV} & \textbf{GLA} & \textbf{GCN} & \textbf{SAGE} & \textbf{GAT} & \textbf{GPS} & \textbf{Sum} & \textbf{Ratio} \\
\midrule
MUTAG & 188 & 0.30 & 0.05 & 0.01 & 0.03 & 0.39 & 1.32 & 1.02 & 2.92 & 4.15 & 9.4 & 23.9$\times$ \\
BZR & 405 & 1.32 & 0.12 & 0.02 & 0.10 & 1.55 & 3.18 & 2.69 & 10.72 & 14.49 & 31.1 & 20.1$\times$ \\
PROTEINS & 1,113 & 4.40 & 0.55 & 0.02 & 0.13 & 5.11 & 10.13 & 7.72 & 36.46 & 82.27 & 136.6 & 26.7$\times$ \\
NCI1 & 4,110 & 12.13 & 3.27 & 0.09 & 3.73 & 19.23 & 32.18 & 28.62 & 137.29 & 152.75 & 350.8 & 18.2$\times$ \\
\bottomrule
\end{tabular}}
\end{sc}
\end{table}

GLA costs 18--27 times less than one training run of all four backbones, and 8--16 times less than GPS alone. Structural summary construction takes most of the GLA budget (63--86\%), while the centroid is nearly free.

These ratios should be read as a measured comparison of these specific runs, not as a general speedup. Two choices pull in opposite directions. On the GLA side, we time one seed, whereas the reference protocol uses five. Multiplying only the probe columns by five gives an estimated ratio of 7--17. On the GNN side, we time a single configuration on a single split, with no evaluation, whereas our grid search (\Cref{app:experiments}) has 12 configurations for GCN and SAGE and 48 for GAT and GPS. A GPU would shorten training and narrow the ratio. Our largest test has 4,110 graphs, and we do not extrapolate beyond it: the probes' share of GLA is largest on NCI1 (36\%), and the RBF-SVM scales superlinearly in the number of graphs (\Cref{app:gla-complexity}), so linear probes may be needed at larger scale.

\subsection{Out-of-Sample Prediction}\label{app:lodo}
The per-backbone predictors in \Cref{app:gla-recipe} are not tested on held-out datasets. We therefore run leave-one-dataset-out cross-validation: a model fitted on 18 benchmarks predicts GCN accuracy on the 19th from its alignment scores alone. A linear model reaches Pearson $r=0.917$, Spearman $\rho=0.821$, MAE 6.30, and RMSE 7.77. Ridge regression gives the same result, and a random forest does worse ($r=0.884$, RMSE 9.18). Since AlignAvg is a linear combination of SLA and FLA, the linear model effectively uses two inputs.

\begin{table}[t]
\centering
\caption{\footnotesize Leave-one-dataset-out prediction of GCN accuracy from SLA, FLA, and AlignAvg with a linear model, on the 19 benchmarks where FLA is defined, sorted by absolute residual (observed minus predicted).}
\label{tab:lodo_gaps}
\footnotesize
\setlength{\tabcolsep}{5pt}
\begin{sc}
\begin{tabular}{lccc|lccc}
\toprule
\textbf{Dataset} & \textbf{Obs.} & \textbf{Pred.} & \textbf{Gap} & \textbf{Dataset} & \textbf{Obs.} & \textbf{Pred.} & \textbf{Gap} \\
\midrule
Synthie & 69.0 & 53.0 & +16.0 & DD & 70.3 & 75.2 & -4.9 \\
COX2 & 78.1 & 65.0 & +13.1 & PROTEINS & 67.6 & 71.0 & -3.4 \\
MOLTOX21 & 73.5 & 86.0 & -12.6 & PROTEINS\_full & 71.1 & 74.3 & -3.3 \\
COLORS-3 & 24.9 & 35.8 & -10.9 & BACE & 72.8 & 70.4 & +2.3 \\
CIFAR-10 & 27.8 & 36.8 & -9.0 & Letter-High & 41.4 & 43.4 & -2.0 \\
MUTAG & 71.3 & 79.7 & -8.4 & ENZYMES & 47.5 & 46.0 & +1.6 \\
AIDS & 98.4 & 90.4 & +8.0 & NCI1 & 68.3 & 67.2 & +1.1 \\
COIL-DEL & 34.0 & 26.2 & +7.8 & MOLHIV & 71.4 & 72.0 & -0.6 \\
BZR & 81.5 & 74.2 & +7.3 & MNIST & 37.9 & 38.0 & -0.1 \\
FRANKENSTEIN & 59.1 & 66.3 & -7.1 & & & & \\
\bottomrule
\end{tabular}
\end{sc}
\end{table}

\Cref{tab:lodo_gaps} reports the residual for every dataset. The eight residuals larger than the RMSE are the cases worth inspecting. Positive gaps (Synthie, COX2, AIDS, COIL-DEL) mark datasets where GCN exceeds what alignment predicts, and negative gaps (MOLTOX21, COLORS-3, CIFAR-10, MUTAG) mark datasets where it falls short. With 19 datasets and a typical error of 6--8 points, this supports ranking datasets and flagging large departures. It does not support precise accuracy estimates, and we have tested it only for GCN. This suggests a three-step workflow: compute SLA and FLA, predict accuracy with a model fitted on other datasets, and spend training budget where the prediction or the SLA--FLA gap warrants it.

\subsection{Benchmark Triage}\label{app:delta-graph}
Following the call to validate graph benchmarks against baselines that ignore relational structure~\citep{bechlerspeicher2025position,coupette2025rings}, we define
\[\Delta_{\mathrm{graph}} = \mathrm{Acc}(\text{best pooled backbone}) - \max(\mathrm{SLA}, \mathrm{FLA}),\]
the accuracy a trained graph model adds over the best no-graph probe. It requires no training beyond the baselines in \Cref{tab:gnn_baselines_extended}. The restricted views correspond to the modes in RINGS: the feature-only view discards the graph and the structure-only view discards features. RINGS estimates these modes by retraining GNNs, whereas we use shallow probes, so $\Delta_{\mathrm{graph}}$ can act as a cheap screen before perturbation-based evaluation. Since our probes are deliberately simple, $\Delta_{\mathrm{graph}}$ is an upper bound on what relational modeling adds under this protocol, and it depends on how well the backbones were tuned. Negative values mean a shallow probe beat every backbone we trained.

\begin{table}[t]
\centering
\caption{\footnotesize $\Delta_{\mathrm{graph}}$ (accuracy points) for the 19 benchmarks where FLA is defined, computed from \Cref{tab:gnn_baselines_extended}.}
\label{tab:delta_graph}
\footnotesize
\setlength{\tabcolsep}{5pt}
\begin{sc}
\begin{tabular}{lr|lr|lr}
\toprule
\multicolumn{2}{c|}{$\Delta_{\mathrm{graph}} \leq +2$} & \multicolumn{2}{c|}{$+2 < \Delta_{\mathrm{graph}} < +10$} & \multicolumn{2}{c}{$\Delta_{\mathrm{graph}} \geq +10$} \\
\midrule
MUTAG & -15.4 & BACE & +5.8 & COX2 & +11.4 \\
MOLTOX21 & -15.0 & BZR & +6.2 & ENZYMES & +14.1 \\
FRANKENSTEIN & -5.9 & & & CIFAR-10 & +17.6 \\
DD & -4.9 & & & MNIST & +23.0 \\
PROTEINS & -4.6 & & & Synthie & +29.7 \\
MOLHIV & -2.2 & & & COIL-DEL & +37.8 \\
AIDS & -0.7 & & & Letter-High & +43.3 \\
PROTEINS\_full & +0.1 & & & & \\
NCI1 & +0.9 & & & & \\
COLORS-3 & +1.9 & & & & \\
\bottomrule
\end{tabular}
\end{sc}
\end{table}


Ten of the 19 benchmarks have $\Delta_{\mathrm{graph}} \leq +2$, and a shallow probe beats the best backbone on seven of them. Nine of the ten are molecular or bioinformatics benchmarks. When accuracy is also high, as on AIDS, these datasets cannot separate methods that use relational structure from methods that discard it, and they should not anchor headline GNN comparisons on their own. Datasets with $\Delta_{\mathrm{graph}} \geq +10$, largely vision-derived or synthetic, are those where trained graph models add accuracy that neither marginal view reaches.

$\Delta_{\mathrm{graph}}$ must be read together with absolute accuracy, because the gap can also vanish when nothing solves the task. COLORS-3 ($\Delta_{\mathrm{graph}}=+1.9$ at 26\% best accuracy) and the Twins are of this kind. They are hard rather than saturated, and they belong in benchmark suites. $\Delta_{\mathrm{graph}}$ is undefined on the featureless social benchmarks.

\subsection{Scaffold Splits}\label{app:scaffold}
Our real-benchmark protocol uses random splits, so it measures in-distribution accessibility. Molecular evaluation often uses scaffold splits, which hold out molecular scaffolds and so test generalization under distribution shift. We recompute SLA and FLA and train a GCN under the official OGB scaffold split, and compare with a random split of the same train and test sizes, which isolates the split type from the split size (\Cref{tab:scaffold}). We report balanced accuracy and ROC-AUC because plain accuracy is uninformative on the imbalanced datasets.

\begin{table}[t]
\centering
\caption{\footnotesize Official OGB scaffold split (S) against a random split of the same train and test sizes (R). SLA and FLA average LogReg and RBF-SVM, and R averages 5 seeds. GCN is a single untuned run on each split with unweighted cross-entropy. BA is balanced accuracy and AUC is ROC-AUC, using the first task on multi-task datasets.}
\label{tab:scaffold}
\footnotesize
\setlength{\tabcolsep}{5pt}
\begin{sc}
\resizebox{\textwidth}{!}{%
\begin{tabular}{l|cccc|cccc|cccc}
\toprule
& \multicolumn{4}{c|}{\textbf{SLA}} & \multicolumn{4}{c|}{\textbf{FLA}} & \multicolumn{4}{c}{\textbf{GCN}} \\
& \multicolumn{2}{c}{BA} & \multicolumn{2}{c|}{AUC} & \multicolumn{2}{c}{BA} & \multicolumn{2}{c|}{AUC} & \multicolumn{2}{c}{BA} & \multicolumn{2}{c}{AUC} \\
\textbf{Dataset} & S & R & S & R & S & R & S & R & S & R & S & R \\
\midrule
BACE & 0.66 & 0.72 & 0.75 & 0.79 & 0.62 & 0.65 & 0.73 & 0.73 & 0.54 & 0.70 & 0.78 & 0.81 \\
MOLTOX21 & 0.64 & 0.73 & 0.74 & 0.79 & 0.70 & 0.73 & 0.80 & 0.80 & 0.68 & 0.58 & 0.76 & 0.72 \\
MOLHIV & 0.65 & 0.65 & 0.70 & 0.71 & 0.63 & 0.63 & 0.64 & 0.67 & 0.50 & 0.50 & 0.62 & 0.69 \\
BBBP & 0.56 & 0.72 & 0.60 & 0.80 & 0.58 & 0.76 & 0.64 & 0.83 & 0.52 & 0.76 & 0.65 & 0.93 \\
ClinTox & 0.49 & 0.52 & 0.53 & 0.62 & 0.54 & 0.66 & 0.58 & 0.71 & 0.50 & 0.50 & 0.61 & 0.75 \\
SIDER & 0.58 & 0.58 & 0.58 & 0.62 & 0.57 & 0.59 & 0.56 & 0.64 & 0.55 & 0.61 & 0.57 & 0.61 \\
\bottomrule
\end{tabular}}
\end{sc}
\end{table}

Scaffold splits change the picture on most of these datasets. By ROC-AUC, GCN falls on five of the six, all except MOLTOX21. SLA falls on all six, though only slightly on MOLHIV, and FLA falls on four, staying flat on BACE and MOLTOX21. The three scores also agree on where the effect is largest: BBBP shows the largest drop for all of them ($-0.20$ for SLA and FLA, $-0.28$ for GCN), followed by ClinTox. Alignment therefore tends to weaken along with GCN under this shift, but it does not track GCN dataset by dataset, as MOLTOX21 shows.

We read these results as a caution about scope. Our random-split correlations do not establish performance under distribution shift, and a practitioner using scaffold splits should compute alignment on those splits. The evidence is also limited. GCN is a single untuned run per split, and its unweighted loss differs from the protocol behind \Cref{tab:gnn_baselines_extended}. On MOLHIV and ClinTox it collapses to the majority class on both splits (balanced accuracy near 0.50), so only ROC-AUC is informative there. Test sets on the smaller datasets have only 143--204 molecules, so small differences should not be over-read.

\section{Limitations} \label{app:limitations}
Our framework is intended as a diagnostic rather than a replacement for full model evaluation. SLA and FLA measure label accessibility under fixed structure-only and feature-only summaries, so their values are relative to the chosen summaries and probes.

\textbf{Summary and probe dependence.} Low alignment on real data may reflect weak marginal signal, interaction-dependent signal, or limits of the summaries and probes. Richer feature summaries and stronger probes raise scores on some low-alignment datasets (\Cref{app:probe-sufficiency,app:fla-ablation}), so such scores are best read as lower bounds. Twins are the exception: because their marginals are matched across classes by construction, no summary of either restricted view can separate them. Rankings and joint regimes are stable across the fixed summaries and probes we tested (\Cref{app:regime-stability}), but we did not evaluate trained summaries such as Deep Sets, and we varied structural blocks without varying the number of eigenvalues or the filtration resolution within a block.

\textbf{Observational and interventional evidence.} Correlations on real benchmarks are descriptive, since domain, size, features, and class count vary together. Evidence that alignment affects learnability comes from the synthetic sweep, where structural and feature separation are set directly while other factors are held fixed, and from Twins.

\textbf{Split protocol.} Our real-benchmark correlations are an in-distribution diagnostic. Seed and split stability holds for random splits (\Cref{app:seed-stability}), while scaffold splits can change the picture (\Cref{app:scaffold}). We evaluated scaffold splits on molecular benchmarks only, and other domain-aware protocols remain open.

\textbf{Scope.} The framework requires a label that belongs to the whole graph, so that discarding edges or features is a well-defined intervention. Our ego-graph experiments recast node classification in this form and show where the diagnostic weakens (\Cref{app:ego_details}). Graph regression should need only regressors in place of the probes, though we have not tested this. Link prediction is harder: its label belongs to a node pair and structure is partly what is being predicted, so its restricted views would need to be redefined. Our study also focuses on standard pooled classifiers and benchmarks of limited size, with screening cost measured up to 4,110 graphs (\Cref{app:runtime}).

These limitations do not affect the main conclusion: graph label alignment provides a lightweight, architecture-agnostic way to organize benchmark difficulty and identify when standard pooled pipelines are likely to succeed or struggle.

\section{Broader Impacts} \label{app:broader}
This paper introduces graph label alignment, a probe based diagnostic that estimates where label signal resides in graph classification datasets by measuring structural and feature marginal accessibility via SLA and FLA. The main positive impact is improved transparency and rigor in benchmark driven graph learning: researchers can better interpret reported gains, identify when high accuracy is likely due to coarse shortcuts, and design evaluations that meaningfully test relational reasoning, which can ultimately improve model reliability in scientific applications such as chemistry and biology. The proposed alignment plane and controlled testbeds may also help practitioners choose appropriate model families and avoid unnecessary compute by revealing when pooled architectures are unlikely to succeed. Potential negative impacts include misuse of these diagnostics to engineer datasets or leaderboards that favor specific methods, or to over simplify conclusions by treating SLA and FLA as definitive measures of task difficulty rather than informative probes. In addition, applying the framework to sensitive relational datasets could facilitate more effective classification of individuals or groups, with attendant privacy and fairness concerns. We encourage responsible use by reporting alignment scores alongside standard metrics, validating conclusions across multiple descriptors and probes, auditing for subgroup harms when labels permit, and avoiding deployment of graph classifiers on high stakes social data without careful governance and monitoring.

\end{document}